\documentclass[11pt]{article}

\newif\ifaclreview
\aclreviewfalse

\IfFileExists{acl.sty}{%
  \ifaclreview\usepackage[review]{acl}\else\usepackage[preprint]{acl}\fi
}{%
  \input{acl_fallback}%
}

\makeatletter
\ifdefined\@LN@makecol
  \IfFormatAtLeastTF{2026-06-01}{}{%
    \RemoveFromHook{build/column/before}[lineno]%
    \AtBeginDocument{%
      \def\@LN@makecol{%
        \@LN@orig@makecol
        \setbox\@outputbox\vbox{%
          \boxmaxdepth\@maxdepth
          \protected@write\@auxout{}{%
            \string\def\string\@LN@column{\if@firstcolumn1\else2\fi}}%
          \box\@outputbox}}%
    }%
  }
\fi
\makeatother

\usepackage{times}
\usepackage{latexsym}
\usepackage[T1]{fontenc}
\usepackage[utf8]{inputenc}
\usepackage{microtype}
\usepackage{inconsolata}
\usepackage{graphicx}
\usepackage{amsmath,amssymb,bm}
\usepackage{amsthm}
\newtheorem{proposition}{Proposition}
\newtheorem{lemma}{Lemma}
\usepackage{booktabs}
\usepackage{enumitem}
\usepackage{multirow}
\usepackage{array}
\usepackage{xcolor}
\usepackage{tikz}
\usepackage{pgfplots}
\pgfplotsset{compat=1.18}
\usepgfplotslibrary{groupplots}
\usetikzlibrary{arrows.meta,calc,decorations.pathreplacing,patterns}

\newcommand{\mthree}{M3}
\newcommand{\mformer}{M3Former}

\definecolor{cq}{RGB}{0,114,178}
\definecolor{ck}{RGB}{230,159,0}
\definecolor{cv}{RGB}{0,158,115}
\definecolor{cr}{RGB}{204,121,167}
\colorlet{cmt}{red!75!black}  %
\colorlet{cb}{black!55}       %

\def\mthrpending{0}

\def\muddequiv{(8.1286e7,1.31438) (3.08332e8,1.20670) (4.58359e8,1.13696) (6.42336e8,1.17022)}

\def\mthrequiv{(8.1286e7,1.33554) (3.08332e8,1.26618) (4.58359e8,1.20718) (6.42336e8,1.21384) (1.36328e9,1.18667)}

\def\attnequiv{(8.1286e7,1.17081) (3.08332e8,1.09426) (4.58359e8,1.00488) (6.42336e8,1.01135) (1.36328e9,1.08341)}

\newcommand\addcoords[2]{\edef\tmp{\noexpand\addplot[\unexpanded{#1}] coordinates {#2};}\tmp}

\title{\mformer{}: Fine-Grained Cross-Layer Mixing for Transformers}

\ifaclreview
\author{Anonymous ACL submission}
\else
\author{%
  Jiasong Han\textsuperscript{1,*},
  Yunkai Chai\textsuperscript{2,3,*},
  Tong Zhu\textsuperscript{3},
  Hengyu Zhang\textsuperscript{1}, \\
  \textbf{Wei Fang,
  Jintao Wang\textsuperscript{1},
  Guoqi Li\textsuperscript{4},
  Yu Cheng\textsuperscript{5,\dag}} \\
  \textmd{\textsuperscript{1}Tsinghua University, Beijing, China} \\
  \textmd{\textsuperscript{2}Shanghai Jiao Tong University, Shanghai, China} \\
  \textmd{\textsuperscript{3}Shanghai AI Laboratory, Shanghai, China} \\
  \textmd{\textsuperscript{4}Institute of Automation, Chinese Academy of Sciences, Beijing, China} \\
  \textmd{\textsuperscript{5}Nanyang Technological University, Singapore} \\
  \textmd{\small\texttt{hanjs22@mails.tsinghua.edu.cn}, \texttt{chaiyunkai@sjtu.edu.cn}, \texttt{zhutong@pjlab.org.cn},} \\
  \textmd{\small\texttt{zhanghen23@mails.tsinghua.edu.cn}, \texttt{fangwei123456g@gmail.com}, \texttt{wangjintao@tsinghua.edu.cn},} \\
  \textmd{\small\texttt{guoqi.li@ia.ac.cn}, \texttt{yu.cheng@ntu.edu.sg}} \\
  \textmd{\small\textsuperscript{*}Equal contribution. \quad \textsuperscript{\dag}Corresponding author.}
}
\fi

\begin{document}
\maketitle

\begin{abstract}
Residual connections give each Transformer block access to all earlier layers only through one fixed sum, broadcast to all of the block's inputs.
This \emph{broadcast bottleneck} imposes three constraints: the query, key, value and residual inputs of a block, which we call its \emph{roles}, read the same vector; all hidden dimensions within a role share one set of weights over earlier layers; and every earlier layer has the same fixed weight for every token.
We propose the Multi-Role Multi-Group Mixing Residual (\mthree{}), which removes all three: every role, and every \emph{group} of hidden dimensions within a role, reads its own token-dependent mixture of all earlier layers.
\mthree{} starts as an exact residual connection and adds as little as 1.4\% parameters and 1.8\% FLOPs.
From 100M to 1.4B parameters, \mformer{} outperforms the Transformer and coarse-grained cross-layer connections at every size, matching a Transformer 1.19--1.34$\times$ the size of the one it modifies, trained with 1.4--1.8$\times$ the compute.
Relaxing only some of the constraints recovers only part of this gain.
\mformer{} also improves sparse mixture-of-experts models of up to 4.6B parameters and trains more stably in deeper networks.
Code: \href{https://github.com/Kingchet10/m3former}{github.com/Kingchet10/m3former}.
\end{abstract}

\section{Introduction}
\label{sec:intro}

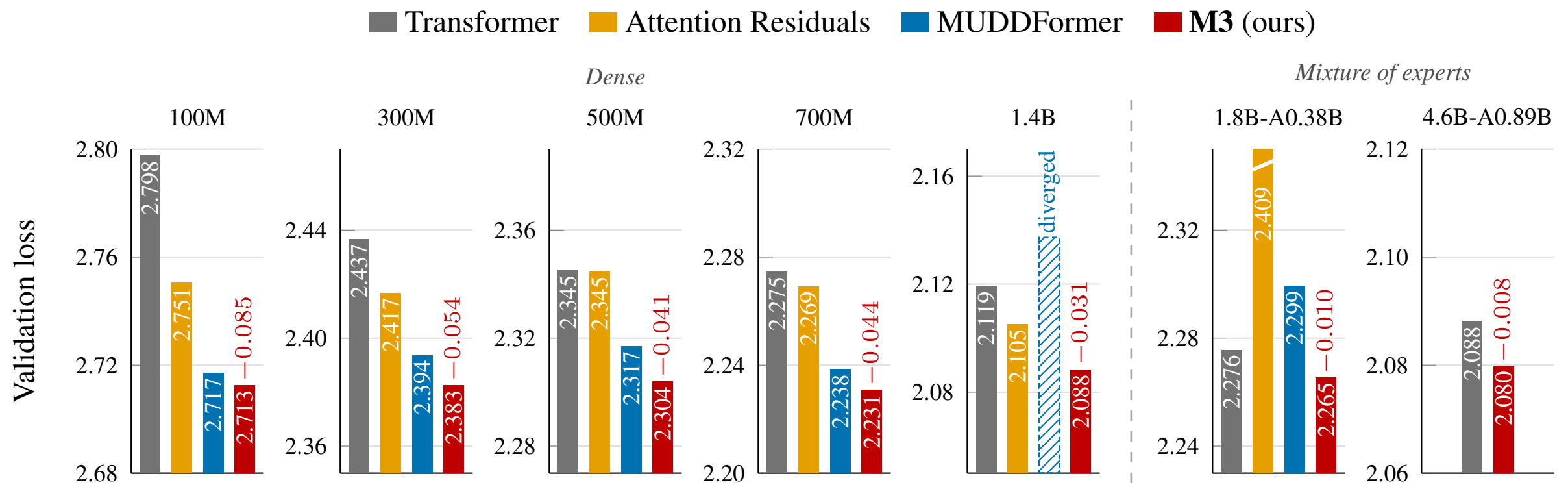
\begin{figure*}[t]
\centering
\begin{tikzpicture}
\pgfplotsset{m3bars/.style={
  width=1.35cm, height=3.3cm, scale only axis,
  xmin=0.4, xmax=4.6, xtick=\empty, scaled y ticks=false,
  tick label style={font=\scriptsize}, title style={font=\scriptsize, yshift=-0.6ex},
  axis x line*=bottom, axis y line*=left,
  ymajorgrids, grid style={black!12},
  clip=false}}
\begin{groupplot}[group style={group name=dense, group size=5 by 1, horizontal sep=0.78cm}, m3bars]
\nextgroupplot[title={100M}, ymin=0, ymax=0.12, ytick={0.0000,0.0400,0.0800,0.1200}, yticklabels={2.68,2.72,2.76,2.80}]
\addplot[ybar, bar width=6pt, bar shift=0pt, fill=cb, draw=none] coordinates {(1,0.11760)};
\node[font=\fontsize{7}{8}\selectfont, rotate=90, anchor=east, text=white, inner sep=1.2pt] at (axis cs:1,0.11760) {2.798};
\addplot[ybar, bar width=6pt, bar shift=0pt, fill=ck, draw=none] coordinates {(2,0.07066)};
\node[font=\fontsize{7}{8}\selectfont, rotate=90, anchor=east, text=white, inner sep=1.2pt] at (axis cs:2,0.07066) {2.751};
\addplot[ybar, bar width=6pt, bar shift=0pt, fill=cq, draw=none] coordinates {(3,0.03719)};
\node[font=\fontsize{7}{8}\selectfont, rotate=90, anchor=east, text=white, inner sep=1.2pt] at (axis cs:3,0.03719) {2.717};
\addplot[ybar, bar width=6pt, bar shift=0pt, fill=cmt, draw=none] coordinates {(4,0.03262)};
\node[font=\fontsize{7}{8}\selectfont, rotate=90, anchor=east, text=white, inner sep=1.2pt] at (axis cs:4,0.03262) {2.713};
\node[font=\fontsize{7}{8}\selectfont\bfseries, text=cmt, rotate=90, anchor=west, inner sep=1pt] at (axis cs:4,0.03262) {$-0.085$};
\nextgroupplot[title={300M}, ymin=0, ymax=0.12, ytick={0.0100,0.0500,0.0900}, yticklabels={2.36,2.40,2.44}]
\addplot[ybar, bar width=6pt, bar shift=0pt, fill=cb, draw=none] coordinates {(1,0.08674)};
\node[font=\fontsize{7}{8}\selectfont, rotate=90, anchor=east, text=white, inner sep=1.2pt] at (axis cs:1,0.08674) {2.437};
\addplot[ybar, bar width=6pt, bar shift=0pt, fill=ck, draw=none] coordinates {(2,0.06677)};
\node[font=\fontsize{7}{8}\selectfont, rotate=90, anchor=east, text=white, inner sep=1.2pt] at (axis cs:2,0.06677) {2.417};
\addplot[ybar, bar width=6pt, bar shift=0pt, fill=cq, draw=none] coordinates {(3,0.04373)};
\node[font=\fontsize{7}{8}\selectfont, rotate=90, anchor=east, text=white, inner sep=1.2pt] at (axis cs:3,0.04373) {2.394};
\addplot[ybar, bar width=6pt, bar shift=0pt, fill=cmt, draw=none] coordinates {(4,0.03253)};
\node[font=\fontsize{7}{8}\selectfont, rotate=90, anchor=east, text=white, inner sep=1.2pt] at (axis cs:4,0.03253) {2.383};
\node[font=\fontsize{7}{8}\selectfont\bfseries, text=cmt, rotate=90, anchor=west, inner sep=1pt] at (axis cs:4,0.03253) {$-0.054$};
\nextgroupplot[title={500M}, ymin=0, ymax=0.12, ytick={0.0100,0.0500,0.0900}, yticklabels={2.28,2.32,2.36}]
\addplot[ybar, bar width=6pt, bar shift=0pt, fill=cb, draw=none] coordinates {(1,0.07516)};
\node[font=\fontsize{7}{8}\selectfont, rotate=90, anchor=east, text=white, inner sep=1.2pt] at (axis cs:1,0.07516) {2.345};
\addplot[ybar, bar width=6pt, bar shift=0pt, fill=ck, draw=none] coordinates {(2,0.07471)};
\node[font=\fontsize{7}{8}\selectfont, rotate=90, anchor=east, text=white, inner sep=1.2pt] at (axis cs:2,0.07471) {2.345};
\addplot[ybar, bar width=6pt, bar shift=0pt, fill=cq, draw=none] coordinates {(3,0.04712)};
\node[font=\fontsize{7}{8}\selectfont, rotate=90, anchor=east, text=white, inner sep=1.2pt] at (axis cs:3,0.04712) {2.317};
\addplot[ybar, bar width=6pt, bar shift=0pt, fill=cmt, draw=none] coordinates {(4,0.03393)};
\node[font=\fontsize{7}{8}\selectfont, rotate=90, anchor=east, text=white, inner sep=1.2pt] at (axis cs:4,0.03393) {2.304};
\node[font=\fontsize{7}{8}\selectfont\bfseries, text=cmt, rotate=90, anchor=west, inner sep=1pt] at (axis cs:4,0.03393) {$-0.041$};
\nextgroupplot[title={700M}, ymin=0, ymax=0.12, ytick={0.0000,0.0400,0.0800,0.1200}, yticklabels={2.20,2.24,2.28,2.32}]
\addplot[ybar, bar width=6pt, bar shift=0pt, fill=cb, draw=none] coordinates {(1,0.07465)};
\node[font=\fontsize{7}{8}\selectfont, rotate=90, anchor=east, text=white, inner sep=1.2pt] at (axis cs:1,0.07465) {2.275};
\addplot[ybar, bar width=6pt, bar shift=0pt, fill=ck, draw=none] coordinates {(2,0.06925)};
\node[font=\fontsize{7}{8}\selectfont, rotate=90, anchor=east, text=white, inner sep=1.2pt] at (axis cs:2,0.06925) {2.269};
\addplot[ybar, bar width=6pt, bar shift=0pt, fill=cq, draw=none] coordinates {(3,0.03849)};
\node[font=\fontsize{7}{8}\selectfont, rotate=90, anchor=east, text=white, inner sep=1.2pt] at (axis cs:3,0.03849) {2.238};
\addplot[ybar, bar width=6pt, bar shift=0pt, fill=cmt, draw=none] coordinates {(4,0.03089)};
\node[font=\fontsize{7}{8}\selectfont, rotate=90, anchor=east, text=white, inner sep=1.2pt] at (axis cs:4,0.03089) {2.231};
\node[font=\fontsize{7}{8}\selectfont\bfseries, text=cmt, rotate=90, anchor=west, inner sep=1pt] at (axis cs:4,0.03089) {$-0.044$};
\nextgroupplot[title={1.4B}, ymin=0, ymax=0.12, ytick={0.0300,0.0700,0.1100}, yticklabels={2.08,2.12,2.16}]
\addplot[ybar, bar width=6pt, bar shift=0pt, fill=cb, draw=none] coordinates {(1,0.06927)};
\node[font=\fontsize{7}{8}\selectfont, rotate=90, anchor=east, text=white, inner sep=1.2pt] at (axis cs:1,0.06927) {2.119};
\addplot[ybar, bar width=6pt, bar shift=0pt, fill=ck, draw=none] coordinates {(2,0.05528)};
\node[font=\fontsize{7}{8}\selectfont, rotate=90, anchor=east, text=white, inner sep=1.2pt] at (axis cs:2,0.05528) {2.105};
\addplot[ybar, bar width=6pt, bar shift=0pt, fill=cmt, draw=none] coordinates {(4,0.03843)};
\node[font=\fontsize{7}{8}\selectfont, rotate=90, anchor=east, text=white, inner sep=1.2pt] at (axis cs:4,0.03843) {2.088};
\node[font=\fontsize{7}{8}\selectfont\bfseries, text=cmt, rotate=90, anchor=west, inner sep=1pt] at (axis cs:4,0.03843) {$-0.031$};
\addplot[ybar, bar width=6pt, bar shift=0pt, draw=cq, line width=0.5pt, dash pattern=on 1.5pt off 1pt, pattern=north east lines, pattern color=cq] coordinates {(3,0.08727)};
\node[font=\fontsize{7}{8}\selectfont, text=cq, rotate=90, anchor=west, inner sep=1pt] at (axis cs:3,0.08727) {diverged};
\end{groupplot}
\begin{groupplot}[group style={group name=moe, group size=2 by 1, horizontal sep=0.78cm}, m3bars]
\nextgroupplot[at={($(dense c5r1.south east)+(1.15cm,0)$)}, anchor=south west, title={1.8B-A0.38B}, ymin=0, ymax=0.12, ytick={0.0100,0.0500,0.0900}, yticklabels={2.24,2.28,2.32}]
\addplot[ybar, bar width=6pt, bar shift=0pt, fill=cb, draw=none] coordinates {(1,0.04557)};
\node[font=\fontsize{7}{8}\selectfont, rotate=90, anchor=east, text=white, inner sep=1.2pt] at (axis cs:1,0.04557) {2.276};
\addplot[ybar, bar width=6pt, bar shift=0pt, fill=ck, draw=none] coordinates {(2,0.12000)};
\node[font=\fontsize{7}{8}\selectfont, rotate=90, anchor=east, text=white, inner sep=1.2pt] at (axis cs:2,0.10800) {2.409};
\draw[white, line width=1.1pt] (axis cs:2,0.1140) ++(-3.5pt,-1.5pt) -- ++(7pt,3pt);
\addplot[ybar, bar width=6pt, bar shift=0pt, fill=cq, draw=none] coordinates {(3,0.06949)};
\node[font=\fontsize{7}{8}\selectfont, rotate=90, anchor=east, text=white, inner sep=1.2pt] at (axis cs:3,0.06949) {2.299};
\addplot[ybar, bar width=6pt, bar shift=0pt, fill=cmt, draw=none] coordinates {(4,0.03541)};
\node[font=\fontsize{7}{8}\selectfont, rotate=90, anchor=east, text=white, inner sep=1.2pt] at (axis cs:4,0.03541) {2.265};
\node[font=\fontsize{7}{8}\selectfont\bfseries, text=cmt, rotate=90, anchor=west, inner sep=1pt] at (axis cs:4,0.03541) {$-0.010$};
\nextgroupplot[title={4.6B-A0.89B}, ymin=0, ymax=0.06, ytick={0.0000,0.0200,0.0400,0.0600}, yticklabels={2.06,2.08,2.10,2.12}]
\addplot[ybar, bar width=6pt, bar shift=0pt, fill=cb, draw=none] coordinates {(2,0.02814)};
\node[font=\fontsize{7}{8}\selectfont, rotate=90, anchor=east, text=white, inner sep=1.2pt] at (axis cs:2,0.02814) {2.088};
\addplot[ybar, bar width=6pt, bar shift=0pt, fill=cmt, draw=none] coordinates {(3,0.01980)};
\node[font=\fontsize{7}{8}\selectfont, rotate=90, anchor=east, text=white, inner sep=1.2pt] at (axis cs:3,0.01980) {2.080};
\node[font=\fontsize{7}{8}\selectfont\bfseries, text=cmt, rotate=90, anchor=west, inner sep=1pt] at (axis cs:3,0.01980) {$-0.008$};
\end{groupplot}
\draw[black!35, dashed, line width=0.5pt] ($(dense c5r1.south east)+(0.32cm,-0.1cm)$) -- ++(0,3.9cm);
\node[anchor=south, font=\scriptsize\itshape, text=black!70] at ($(dense c1r1.north west)!0.5!(dense c5r1.north east)+(0,0.52cm)$) {Dense};
\node[anchor=south, font=\scriptsize\itshape, text=black!70] at ($(moe c1r1.north west)!0.5!(moe c2r1.north east)+(0,0.52cm)$) {Mixture of experts};
\node[anchor=south, font=\small] at ($(dense c1r1.north west)!0.5!(moe c2r1.north east)+(0,1.0cm)$) {%
  \tikz\fill[cb] (0,0) rectangle (0.28,0.2); Transformer\quad
  \tikz\fill[ck] (0,0) rectangle (0.28,0.2); Attention Residuals\quad
  \tikz\fill[cq] (0,0) rectangle (0.28,0.2); MUDDFormer\quad
  \tikz\fill[cmt] (0,0) rectangle (0.28,0.2); \textbf{M3} (ours)};
\node[anchor=south, rotate=90, font=\small] at ($(dense c1r1.west)+(-0.85cm,0)$) {Validation loss};
\end{tikzpicture}
\caption{Validation loss at 20 training tokens per parameter (active parameters for mixture-of-experts models). In the bars: losses; red: change of \mformer{} relative to the Transformer; hatched: diverged run; white slash: bar cut off at the top. Each panel spans 0.12 in loss, the last 0.06.}
\label{fig:lossbars}
\end{figure*}

Residual connections \citep{he2016deep} allow Transformers \citep{vaswani2017attention} to be trained at great depth.
Each block adds its output to a running sum, and the next block reads this sum as its input.
In a pre-norm Transformer, the input of a block is thus the token embedding plus the updates of all earlier blocks.
This running sum, often called the residual stream, is the only path through which a block can use what earlier layers have computed \citep{elhage2021mathematical}.
Although language models have grown to billions of parameters \citep{brown2020language,kaplan2020scaling,hoffmann2022training}, nearly all of them still connect their blocks in this way.

\begin{figure*}[!t]
\centering
\resizebox{\textwidth}{!}{\input{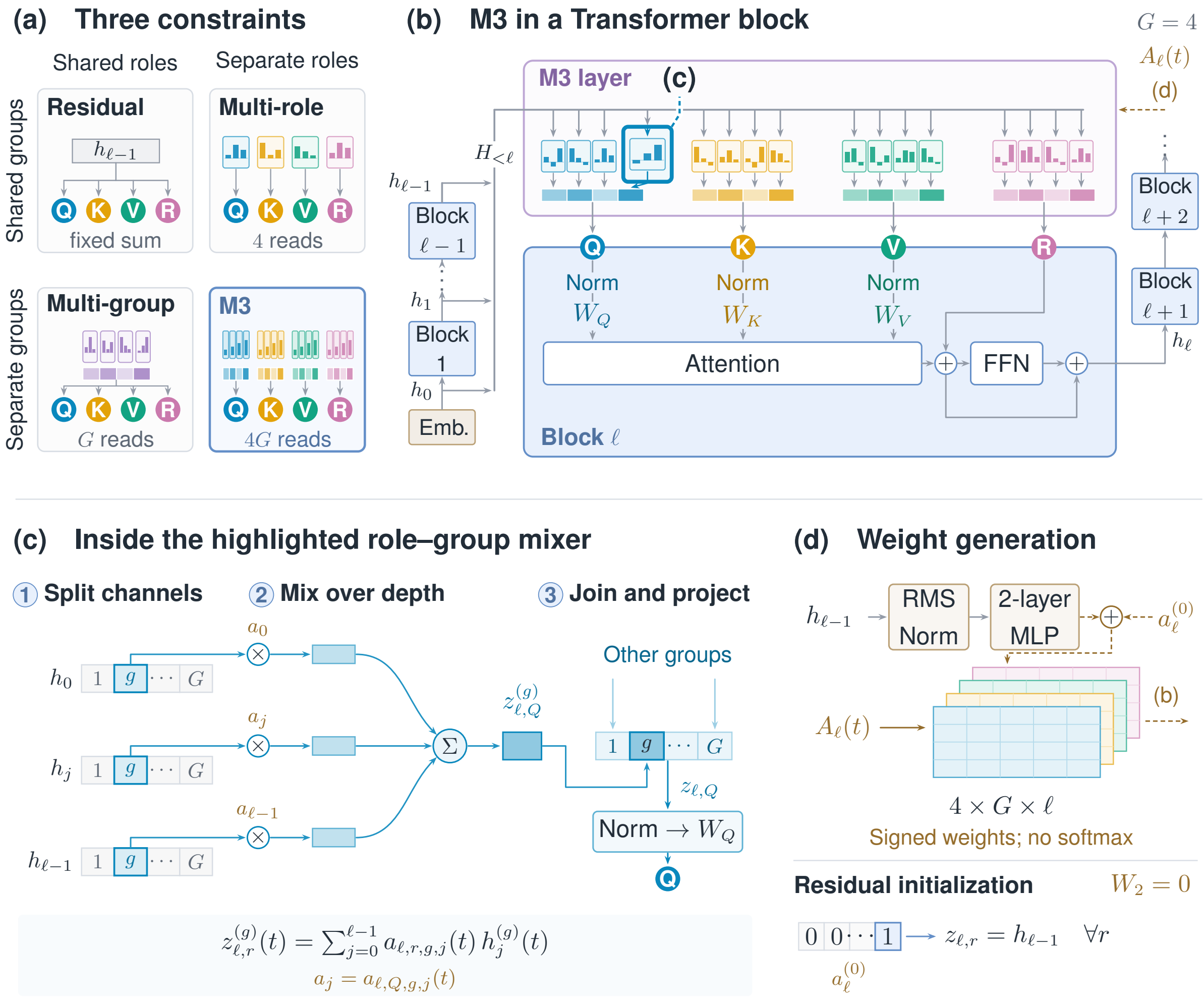}}
\caption{\mthree{}: multi-role multi-group mixing.
(a)~The three constraints: a residual connection gives every role and group the same fixed sum $h_{\ell-1}$, whereas \mthree{} gives each of the $4G$ role--group pairs its own token-dependent mixture.
(b)~Block~$\ell$ assembles each of its four role inputs from $G$ group-wise mixtures of the history $H_{<\ell}$.
(c)~A role--group mixer weights group $g$ of every earlier state by $a_{\ell,r,g,j}(t)$ and sums them.
(d)~A generator computes all coefficients from $h_{\ell-1}(t)$; its initialization reproduces the standard Transformer.
$G=4$ is shown; coefficient profiles are schematic.}
\label{fig:sharing}
\end{figure*}

Every block reads only this sum and broadcasts it to all of its inputs, although the readers of the stream need different information.
The broadcast constrains a block in three ways (Figure~\ref{fig:sharing}a).
(1)~\textbf{Roles}: queries and keys decide where to attend, values decide what to move, and the residual path carries the state that the block updates \citep{vaswani2017attention,elhage2021mathematical}, yet all four read the same vector.
(2)~\textbf{Groups}: different subspaces of the stream carry different features \citep{elhage2021mathematical}, and heads and layers specialize in different functions \citep{voita2019analyzing,clark2019does,tenney2019bert,jawahar2019does}, yet all hidden dimensions of a role share one set of weights over earlier layers.
(3)~\textbf{Layer weights}: every earlier layer enters the sum with the same fixed weight for every token, so a block cannot emphasize the layers it needs; in pre-norm models the growing sum dilutes each layer \citep{kimi2026attnres}, and many deep layers can be removed with little loss \citep{gromov2024unreasonable}.
We call these constraints the \emph{broadcast bottleneck}.

Cross-layer connections relax these constraints one at a time, and each relaxation has improved language modeling.
Most relax~(3), with static \citep{wang2019learning,pagliardini2024denseformer} or token-dependent weights over earlier layers \citep{zhu2025hyperconnections,xie2025mhc,kimi2026attnres}; some also relax~(1) by forming the query, key and value from separate mixtures \citep{heddes2025deepcrossattention,xiao2025muddformer}, or~(2) by splitting the hidden state into fractions, heads or streams \citep{zhu2025frac,luo2026mhar,qin2026msar}.
None relaxes all three (Table~\ref{tab:related}).
Yet every group of every role needs different information from different earlier layers, and this need changes from token to token; keeping any one constraint still forces some of them to share a mixture, so removing the bottleneck requires relaxing all three together.

In this work, we propose the \textbf{Multi-Role Multi-Group Mixing Residual} (\mthree{}), which relaxes all three constraints jointly: every role of a block, and every group of hidden dimensions within a role, combines all earlier layers with its own weights, generated from the current token (Figure~\ref{fig:sharing}).
Because it starts as an exact residual connection, \mthree{} can replace the residual connections of a Transformer without other changes; we call the result \mformer{}.

We pre-train \mformer{}, the Transformer and coarse-grained cross-layer connections from 100M to 1.4B parameters under one recipe, together with mixture-of-experts models of up to 4.6B parameters, and make three contributions:
\begin{itemize}[leftmargin=*,itemsep=2pt,topsep=3pt]
  \item \textbf{Problem.} We identify the broadcast bottleneck, three constraints that the residual connection places on how a block reads earlier layers, and show that each of them limits the Transformer: relaxing only one or two recovers only part of the gain (\S\ref{sec:axes}).
  \item \textbf{Method.} We propose \mthree{}, in which every role--group pair of a block reads its own token-dependent mixture of all earlier layers. One small generator produces all $4G$ mixtures, and a fused kernel computes them in a single pass over the history. The resulting \mformer{} starts exactly as the standard Transformer and adds as little as 1.4\% parameters and 1.8\% FLOPs.
  \item \textbf{Evidence at scale.} At 20 tokens per parameter, \mformer{} matches a Transformer 1.19--1.34$\times$ the size of the one it modifies (Figure~\ref{fig:teaser}) and outperforms attention residuals from 100M to 1.4B and MUDDFormer from 100M to 700M (Figure~\ref{fig:lossbars}). It also improves mixture-of-experts models of up to 4.6B parameters, and it trains to completion at 1.4B, where MUDDFormer diverges under the same recipe, and stays ahead when both use PrePostNorm.
\end{itemize}

\section{\mthree{}: Multi-Role Multi-Group Mixing}
\label{sec:method}

We first write the residual connection as a read of a history of hidden states (\S\ref{sec:history}), then define depth mixing and its granularity (\S\ref{sec:granularity}), describe how \mthree{} computes and initializes its mixing coefficients (\S\ref{sec:weights}), and finally analyze its cost (\S\ref{sec:complexity}).

\subsection{The residual connection reads a history}
\label{sec:history}

Consider a pre-norm Transformer with $L$ blocks and hidden size $d$.
Let $h_0$ be the token embedding and $h_\ell$ the hidden state after block~$\ell$; all quantities are per token, and we write the token index $t$ only where it matters.
Block~$\ell$ has access to the history $H_{<\ell}=(h_0,\dots,h_{\ell-1})$, but a standard block reads only its last element:
\begin{equation}
\begin{aligned}
u_\ell &= h_{\ell-1} + \mathrm{Attn}\big(\hat h W_Q,\ \hat h W_K,\ \hat h W_V\big),\\
h_\ell &= u_\ell + \mathrm{FFN}\big(\mathrm{N}_F(u_\ell)\big),
\end{aligned}\label{eq:block}
\end{equation}
where $\hat h=\mathrm{N}_A(h_{\ell-1})$, $\mathrm{N}_A$ and $\mathrm{N}_F$ are RMSNorms, and $W_Q,W_K,W_V$ are the query, key and value projections.
Because $h_{\ell-1}=h_0+\sum_{k=1}^{\ell-1}(h_k-h_{k-1})$, reading the latest state is the same as adding the updates of all earlier blocks with equal, fixed weights.
The block uses this one vector in four \emph{roles}: as the input of the query, key and value projections, and as the residual path to which the attention output is added.

\mthree{} gives the four roles separate inputs $z_{\ell,r}$, $r\in\mathcal{R}=\{Q,K,V,R\}$:
\begin{equation}
\begin{aligned}
u_\ell &= z_{\ell,R} + \mathrm{Attn}\big(\hat z_Q W_Q,\ \hat z_K W_K,\ \hat z_V W_V\big),\\
h_\ell &= u_\ell + \mathrm{FFN}\big(\mathrm{N}_F(u_\ell)\big),
\end{aligned}\label{eq:m3block}
\end{equation}
where $\hat z_Q=\mathrm{N}_A(z_{\ell,Q})$ reuses the attention norm of Eq.~\eqref{eq:block}, while $\hat z_K=\mathrm{N}_K(z_{\ell,K})$ and $\hat z_V=\mathrm{N}_V(z_{\ell,V})$ use two new RMSNorms, so that each projected role has its own scale.
The residual input $z_{\ell,R}$ is added without normalization, and the feed-forward step is unchanged.
With $z_{\ell,r}=h_{\ell-1}$ for every role and the scales of $\mathrm{N}_K$ and $\mathrm{N}_V$ equal to that of $\mathrm{N}_A$, Eq.~\eqref{eq:m3block} reduces to Eq.~\eqref{eq:block}.
Block~1 has only $h_0$ to read and is left unchanged.

\subsection{Depth mixing and its granularity}
\label{sec:granularity}

We split the channels of every hidden state into $G$ contiguous groups of width $d/G$, $h_j=[h^{(1)}_j;\dots;h^{(G)}_j]$, and form each group of each role input as a weighted sum over the history (Figure~\ref{fig:sharing}c):
\begin{equation}
z^{(g)}_{\ell,r}(t)=\sum_{j=0}^{\ell-1}a_{\ell,r,g,j}(t)\,h^{(g)}_j(t).\label{eq:mix}
\end{equation}
The groups partition the hidden channels before the query, key and value projections, so they are not attention heads: a role input is reassembled from its $G$ groups and only then normalized and projected as in Eq.~\eqref{eq:m3block}.

Eq.~\eqref{eq:mix} defines a family of connections that relax the three constraints of \S\ref{sec:intro} to different degrees (Figure~\ref{fig:sharing}a):
\begin{itemize}[leftmargin=*,itemsep=1pt,topsep=3pt]
  \item \textbf{Residual connection}: $a_{\ell,r,g,j}=\mathbf{1}[j=\ell-1]$, one fixed mixture for all roles and groups; relaxes none.
  \item \textbf{Dense mixing}: $a_{\ell,r,g,j}=a_{\ell,j}$, one mixture of all earlier layers shared by all roles and groups; relaxes~(3). Its weights are static in DenseFormer \citep{pagliardini2024denseformer} and token-dependent in attention residuals \citep{kimi2026attnres}.
  \item \textbf{Multi-role}: $a_{\ell,r,g,j}=a_{\ell,r,j}$, four mixtures, one per role, as in MUDDFormer \citep{xiao2025muddformer}; relaxes (1) and (3).
  \item \textbf{Multi-group}: $a_{\ell,r,g,j}=a_{\ell,g,j}$, $G$ mixtures shared by all roles, as in multi-head attention residuals \citep{luo2026mhar}; relaxes (2) and (3).
  \item \textbf{\mthree{}}: no sharing, $4G$ mixtures, one for every role and group; relaxes all three.
\end{itemize}
We call the number of independent mixtures per block and token the \emph{granularity} of a connection.
The family is nested: \mthree{} represents every coarser member exactly (Proposition~\ref{prop:nest}), and readers that share a coefficient receive the sum of their gradients for it, which cancel in part when they need different sources.
With token-dependent coefficients, existing cross-layer connections relax at most two of the three constraints; DeepCrossAttention \citep{heddes2025deepcrossattention} makes only its token-independent coefficients specific to channels. Table~\ref{tab:related} compares them on all three, and \S\ref{sec:related} discusses them.

\begin{table}[t]
\centering
\small
\setlength{\tabcolsep}{3pt}
\begin{tabular}{@{}lcccc@{}}
\toprule
& (1) & (2) & (3) & \\
Method & Roles & Groups & Weights & Sources \\
\midrule
Residual & -- & -- & fixed & $h_{\ell-1}$ \\
DLCL, DenseFormer & -- & -- & static & all \\
Hyper-conn., mHC & -- & -- & dynamic & streams \\
Frac-connections & -- & \checkmark & dynamic & $h_{\ell-1}$ \\
AttnRes & -- & -- & dynamic & blocks \\
Multi-head AttnRes & -- & \checkmark & dynamic & all \\
Value residual & V & -- & static & layer 1 \\
DeepCrossAttn. & Q,K,V & static & both & all \\
MUDDFormer & Q,K,V,R & -- & dynamic & all \\
\midrule
\mthree{} (ours) & Q,K,V,R & \checkmark & dynamic & all \\
\bottomrule
\end{tabular}
\caption{Cross-layer connections and the three constraints. Static weights are learned but the same for every token; dynamic weights depend on the token.}
\label{tab:related}
\end{table}
\mthree{} relaxes all three: in it, every coefficient is specific to a role and a group and depends on the token (\S\ref{sec:weights}).

\subsection{Token-conditioned weights}
\label{sec:weights}

A single generator per block produces all coefficients of Eq.~\eqref{eq:mix} from the latest state of the same token (Figure~\ref{fig:sharing}d):
\begin{equation}
\begin{aligned}
\mathrm{vec}(A_\ell)&=\mathrm{GELU}\big(\bar{\mathrm{N}}(h_{\ell-1})\,W_{\ell,1}\big)\,W_{\ell,2}\\
&\quad+\mathrm{vec}\big(a^{(0)}_\ell\big),
\end{aligned}\label{eq:gen}
\end{equation}
computed separately for every token, where $A_\ell\in\mathbb{R}^{4\times G\times\ell}$ collects the coefficients $a_{\ell,r,g,j}$, $\mathrm{vec}$ flattens it into a row vector of length $4G\ell$, $\bar{\mathrm{N}}$ is an RMSNorm without a learned scale, $W_{\ell,1}\in\mathbb{R}^{d\times m_\ell}$, $W_{\ell,2}\in\mathbb{R}^{m_\ell\times 4G\ell}$, and $a^{(0)}_\ell\in\mathbb{R}^{4\times G\times\ell}$ is a learned static prior.
The hidden width $m_\ell$ is the smallest multiple of 64 that exceeds $4\ell$, so the generator is small.
The coefficients are signed and are not normalized over depth: a role can subtract an earlier representation as easily as it adds one, which a convex mixture cannot do (Lemma~\ref{lem:updates} in Appendix~\ref{app:derivations}).
The hidden state that enters the output layer is mixed in the same way from $(h_0,\dots,h_L)$, with a single role and its own generator.

\paragraph{Initialization.}
We set $W_{\ell,2}=0$, draw $W_{\ell,1}$ from $\mathcal{N}(0,1/d)$, and set $a^{(0)}_\ell$ to one on the latest state and zero elsewhere.
Then $z_{\ell,r}=h_{\ell-1}$ for every role, and \mformer{} computes exactly the same function as the standard Transformer at the start of training (Proposition~\ref{prop:init}); every departure from a single shared read is learned.

\paragraph{Number of groups.}
We count $G$ in units of the number of attention heads, $H=d/64$: from 100M to 700M we train $G\in\{H,2H,4H\}$, that is, groups of width 64, 32 and 16, and at 1.4B $G=H$; we report the best at each size (\S\ref{sec:groups}, Table~\ref{tab:parambound}).
The groups are not tied to the heads, since they partition the channels of the history before the projections.

\subsection{Complexity analysis}
\label{sec:complexity}

Per token, block~$\ell$ spends $(d+4G\ell)\,m_\ell$ multiply--adds in the generator, equal to the number of entries of its weight matrices, and $4\ell d$ in the mixing of Eq.~\eqref{eq:mix}.
Finer granularity is cheap for two reasons.
First, a group-wise mixture touches the same $\ell d$ values as a full-width one, so \mthree{} spends the same $4\ell d$ multiply--adds on mixing as multi-role mixing; its $4G$ mixtures cost more only in the output layer of the generator, $4G\ell\,m_\ell$ instead of $4\ell\,m_\ell$.
Second, every member of the family reads the same $\ell$ earlier states, and our fused Triton kernel reads each $h_j$ once and writes all four roles and $G$ groups in one pass, so its reads and writes of hidden states do not grow with $G$.

Summed over the network, \mthree{} adds a fraction $O(L/d+GL^2/d^2)$ of the non-embedding parameters of the Transformer, which grows with depth relative to width (Appendix~\ref{app:cost}).
For the models we report, it adds 1.4--3.9\% parameters and 1.8--4.3\% multiply--adds per token (Table~\ref{tab:parambound}), against 0.6--1.0\% and 1.0--1.4\% for MUDDFormer.
At inference, a token keeps only its own $L+1$ hidden states, so the extra memory does not grow with the context length.

\section{Experimental Setup}
\label{sec:setup}

\paragraph{Data and training.}
We train all models on a fixed sample of FineWeb-Edu \citep{penedo2024fineweb}, tokenized with the 32K-token TinyLlama tokenizer \citep{zhang2024tinyllama}, with sequences of 2,048 tokens and a global batch of 64 sequences.
All models use AdamW with a warmup--stable--decay schedule \citep{hu2024minicpm} and one preset learning rate per size, shared by all methods (Appendix~\ref{app:training}).
The training budget of each size is set by the non-embedding parameter count $N$ of its Transformer.
The main run trains on $20N$ tokens, close to the compute-optimal ratio of \citet{hoffmann2022training}, and runs with $5N$ and $10N$ tokens decay from checkpoints of the main run.
We report the loss on 99.9M held-out tokens.

\paragraph{Models.}
Table~\ref{tab:models} lists the five dense sizes and two mixture-of-experts models.
The Transformer is a pre-norm decoder with RMSNorm \citep{zhang2019rmsnorm}, SwiGLU feed-forward layers \citep{shazeer2020glu} and rotary position embeddings \citep{su2024roformer} (Appendix~\ref{app:training}).
Where we report one \mformer{} per size, we use its best $G$.
In the mixture-of-experts models, every feed-forward layer routes each token to 8 of 64 SwiGLU experts, without a shared expert (Table~\ref{tab:moearch}); they also train on 20 tokens per parameter, counting active parameters, and \mthree{} uses $G=4H$ in both, 64 and 96 groups.

\begin{table}[t]
\centering
\small
\setlength{\tabcolsep}{3pt}
\begin{tabular}{@{}lccr@{}}
\toprule
Model & $L\times d$ & FFN & $N$ \\
\midrule
100M & $16\times 640$  & 1792 & 81M   \\
300M & $24\times 1024$ & 2816 & 308M  \\
500M & $28\times 1152$ & 3200 & 458M  \\
700M & $32\times 1280$ & 3520 & 642M  \\
1.4B & $40\times 1664$ & 4608 & 1363M \\
\midrule
1.8B-A0.38B & $24\times 1024$ & $64\times352$ & 1.76B/0.31B \\
4.6B-A0.89B & $28\times 1536$ & $64\times512$ & 4.49B/0.80B \\
\bottomrule
\end{tabular}
\caption{Models. $N$ is the non-embedding parameter count of the Transformer, total/active for the mixture-of-experts models, whose FFN column gives the number and width of experts. All heads have width 64.}
\label{tab:models}
\end{table}

\paragraph{Baselines.}
We compare \mformer{} with two coarse-grained cross-layer connections.
\emph{MUDDFormer} \citep{xiao2025muddformer} lets each role read its own token-dependent mixture of earlier layers, shared by all of its channels; it is \mthree{} with $G=1$, with the same block and normalization.
We train it with the same pre-norm recipe as all other models.
\emph{Attention residuals} \citep{kimi2026attnres} replace the residual sum with softmax attention over the outputs of earlier layers; we train their block variant, with blocks of four layers, from 100M to 1.4B.
For the ablation in \S\ref{sec:axes}, we also train \emph{multi-group mixing}, in which each group reads its own mixture, shared by all roles.

\paragraph{Equivalent size.}
To express a loss as a model size, we fit $L(N)=E+A\,(N/10^8)^{-\alpha}$ to the five Transformers ($E=0.965$, $A=1.771$, $\alpha=0.164$; RMSE 0.002).
The \emph{equivalent size} of a model is $k=N_{\mathrm{eq}}/N$, where $N$ is the size of the Transformer that the model modifies (Table~\ref{tab:models}) and $N_{\mathrm{eq}}$ the size of the Transformer that this fit predicts to reach the same loss; since that Transformer is also trained on 20 tokens per parameter, it uses $k^2$ times the training compute (Appendix~\ref{app:scaling}).

\section{Results}
\label{sec:results}

\subsection{Scaling with model size}
\label{sec:size}

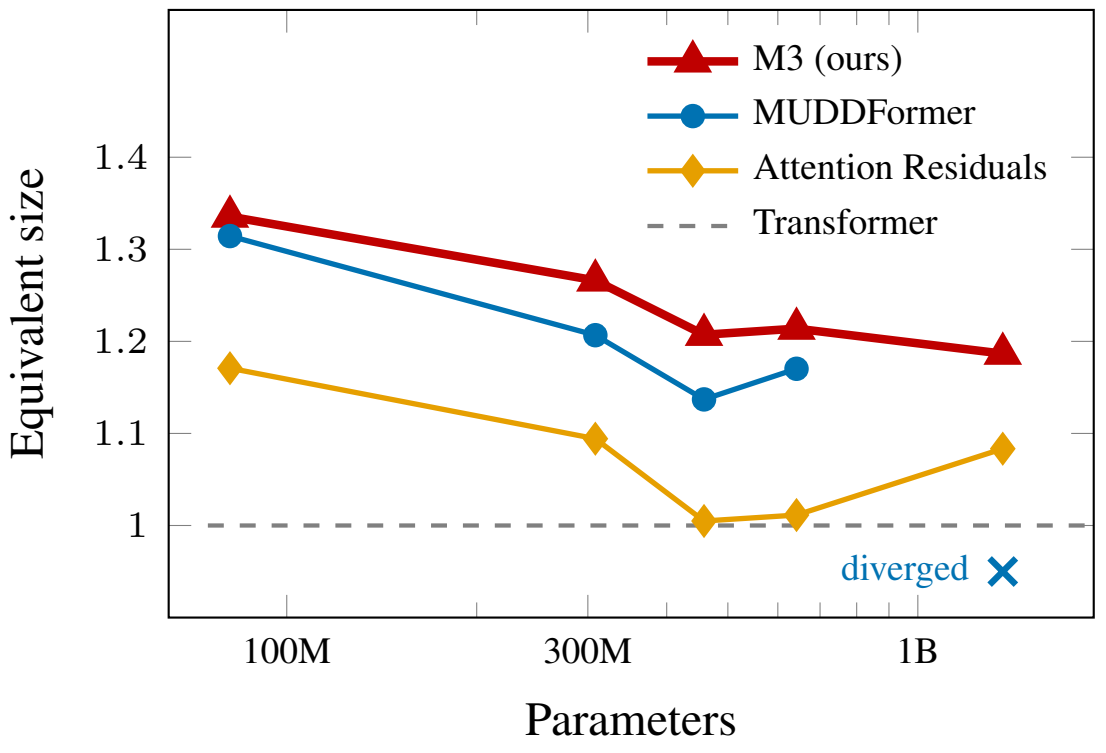
\begin{figure}[t]
\centering
\begin{tikzpicture}
\begin{axis}[
  width=\columnwidth, height=5.6cm,
  xmode=log,
  xmin=6.5e7, xmax=1.9e9,
  ymin=0.9, ymax=1.56,
  xtick={1e8,3e8,1e9},
  xticklabels={100M,300M,1B},
  minor xtick={2e8,4e8,5e8,6e8,7e8,8e8,9e8},
  ytick={1.0,1.1,1.2,1.3,1.4},
  yticklabel style={/pgf/number format/fixed,/pgf/number format/precision=1},
  xlabel={Parameters},
  ylabel={Equivalent size},
  label style={font=\small},
  tick label style={font=\scriptsize},
  legend style={font=\scriptsize,draw=none,fill=none,at={(0.98,0.98)},anchor=north east},
  legend cell align=left,
  every axis plot/.append style={line width=1pt, mark size=1.8pt},
]
\addcoords{color=red!75!black, mark=triangle*, line width=1.6pt, mark size=2.6pt}{\mthrequiv}
\addlegendentry{\mthree{} (ours)}
\addcoords{color=cq, mark=*}{\muddequiv}
\addlegendentry{MUDDFormer}
\addcoords{color=ck, mark=diamond*, mark size=2.4pt}{\attnequiv}
\addlegendentry{Attention Residuals}
\addplot[color=gray, dashed, line width=0.8pt, domain=7.5e7:2.1e9, samples=2] {1};
\addlegendentry{Transformer}
\addplot[color=cq, only marks, mark=x, mark size=3.5pt, line width=1.2pt, forget plot] coordinates {(1363.284e6,0.95)};
\node[font=\scriptsize, text=cq, anchor=east] at (axis cs:1.3e9,0.95) {diverged};
\end{axis}
\end{tikzpicture}
\caption{Equivalent size $k=N_{\mathrm{eq}}/N$ (\S\ref{sec:setup}) at 20 tokens per parameter, against the size $N$ of the Transformer that each model modifies.}
\label{fig:teaser}
\end{figure}

Figure~\ref{fig:lossbars} compares the four models at 20 tokens per parameter, and \mformer{} reaches the lowest loss at every size.
It lowers the loss of the Transformer by 0.085, 0.054, 0.041, 0.044 and 0.031 from 100M to 1.4B, 6--47\% more than MUDDFormer up to 700M; at 1.4B, MUDDFormer diverged, and with PrePostNorm for both, \mformer{} stays 0.010 ahead (\S\ref{sec:stability}).
In equivalent size (Figure~\ref{fig:teaser}), \mformer{} matches a Transformer 1.19--1.34$\times$ the size of the one it modifies, or 1.41--1.78$\times$ its training compute, against 1.14--1.31$\times$ for MUDDFormer.
Spent on a larger Transformer, its 1.4--3.9\% extra parameters would lower the loss by only 0.003--0.008 according to the fit (Appendix~\ref{app:scaling}).
Attention residuals help most at 100M (0.047, or 1.17$\times$), but by at most 0.020 from 300M to 1.4B (1.00--1.09$\times$).

\subsection{Scaling with data}
\label{sec:data}

\begin{figure}[t]
\centering
\definecolor{sz1}{HTML}{F0A030}
\definecolor{sz3}{HTML}{E16462}
\definecolor{sz5}{HTML}{B12A90}
\definecolor{sz7}{HTML}{6A00A8}
\definecolor{sz14}{HTML}{0D0887}
\begin{tikzpicture}
\begin{axis}[
  width=\columnwidth, height=7cm,
  xmode=log, xmin=3e8, xmax=3.6e10,
  xtick={1e9,1e10}, xticklabels={1B,10B},
  minor xtick={4e8,5e8,6e8,7e8,8e8,9e8,2e9,3e9,4e9,5e9,6e9,7e9,8e9,9e9,2e10,3e10},
  ymin=2.03, ymax=3.14,
  ytick={2.2,2.4,2.6,2.8,3.0},
  yticklabel style={/pgf/number format/fixed,/pgf/number format/precision=1},
  xlabel={Training tokens}, ylabel={Validation loss},
  label style={font=\small}, tick label style={font=\scriptsize},
  legend style={font=\scriptsize, draw=none, fill=none, at={(0.99,0.99)}, anchor=north east,
    row sep=-1pt},
  legend cell align=left,
  clip=false,
]
\addlegendimage{black!70, dashed, line width=0.9pt, mark=o, mark options={solid}, mark size=1.6pt}
\addlegendentry{Transformer}
\addlegendimage{black!70, line width=1.2pt, mark=triangle*, mark size=2.2pt}
\addlegendentry{\textbf{M3} (ours)}
\addlegendimage{empty legend}\addlegendentry{}
\addlegendimage{sz1, line width=2pt}\addlegendentry{100M}
\addlegendimage{sz3, line width=2pt}\addlegendentry{300M}
\addlegendimage{sz5, line width=2pt}\addlegendentry{500M}
\addlegendimage{sz7, line width=2pt}\addlegendentry{700M}
\addlegendimage{sz14, line width=2pt}\addlegendentry{1.4B}
\def\tokpair#1#2{%
  \addcoords{#2, dashed, line width=0.9pt, mark=o, mark options={solid}, mark size=1.6pt, forget plot}{\csname basetok#1\endcsname}%
  \expandafter\ifx\csname mthrtok#1\endcsname\empty\else
    \addcoords{#2, line width=1.2pt, mark=triangle*, mark size=2.2pt, forget plot}{\csname mthrtok#1\endcsname}%
  \fi}
\tokpair{one}{sz1}\tokpair{three}{sz3}\tokpair{five}{sz5}\tokpair{seven}{sz7}\tokpair{fourteen}{sz14}
\ifnum\mthrpending=1
  \node[font=\scriptsize, sz14, anchor=center] at (axis cs:1.05e10,2.09) {M3 1.4B: training};
\fi
\end{axis}
\end{tikzpicture}
\caption{Validation loss against training tokens, at 5, 10 and 20 tokens per parameter for every size.
\mthree{} uses the best $G$ of each size.}
\label{fig:data}
\end{figure}
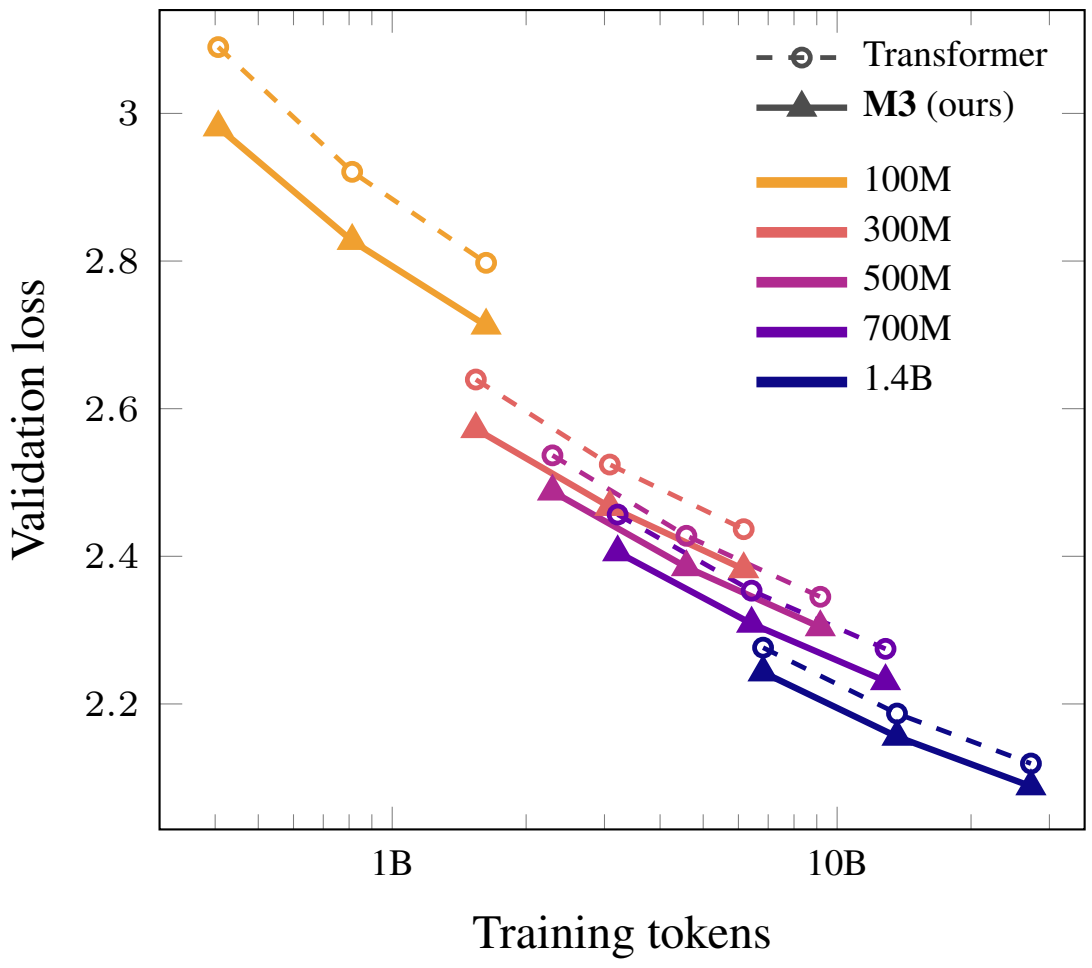

Figure~\ref{fig:data} follows each model as its training data grow from 5 to 20 tokens per parameter.
The curve of \mformer{} lies below that of the Transformer at every size and budget, by 0.031--0.109.
The gain is not just a faster start: from 300M on, it shrinks by at most 0.013 between 5 and 20 tokens per parameter, while the loss itself falls by 0.16--0.20.

\subsection{Mixture-of-experts models}
\label{sec:moe}

Fine-grained mixing also helps sparse models.
On the two mixture-of-experts models of Table~\ref{tab:models}, \mthree{} lowers the loss by 0.010 and 0.008 (Figure~\ref{fig:lossbars}, right).
In the small model, MUDDFormer instead raises the loss by 0.024, and block attention residuals by 0.133.
Both use the dense recipe without tuning, whereas \citet{kimi2026attnres} report gains for attention residuals with their own mixture-of-experts recipe.

\subsection{Downstream evaluation}
\label{sec:downstream}

We evaluate zero-shot and five-shot accuracy with the LM Evaluation Harness \citep{gao2024harness} on the tasks used by \citet{xiao2025muddformer}: LAMBADA \citep{paperno2016lambada}, PIQA \citep{bisk2020piqa}, WinoGrande \citep{sakaguchi2020winogrande}, ARC-Easy and ARC-Challenge \citep{clark2018think}, SciQ \citep{welbl2017sciq}, LogiQA \citep{liu2020logiqa}, BoolQ \citep{clark2019boolq}, HellaSwag \citep{zellers2019hellaswag}, and the middle-school part of RACE \citep{lai2017race}.
Table~\ref{tab:downstream} reports zero-shot accuracy of the 700M and 1.4B dense models and both mixture-of-experts models, and Table~\ref{tab:downstream5} in Appendix~\ref{app:full} five-shot accuracy.
\mthree{} has the highest average zero-shot accuracy in all four models, 0.4--1.3 points above the Transformer and 0.3--0.6 above the strongest other connection, and beats the Transformer on 30 of the 40 model--task pairs.
Its largest and most consistent gain is on LAMBADA, 2.5--3.7 points.
With five examples, \mthree{} has the highest average at 700M, 1.4B and in the large mixture-of-experts model, 0.4--0.7 points above the Transformer, and is 0.3 below it in the small one.

\begin{table*}[t]
\centering
\scriptsize
\setlength{\tabcolsep}{4.2pt}
\begin{tabular}{@{}llcccccccccccc@{}}
\toprule
Size & Model & PPL$\downarrow$ & LAMBADA & PIQA & WinoGrande & ARC-E & ARC-C & SciQ & LogiQA & BoolQ & HellaSwag & RACE-M & Avg \\
\midrule
\multirow{4}{*}{700M} & Transformer & 9.72 & 30.60 & 68.61 & 54.30 & \textbf{60.31} & 29.78 & 88.30 & 23.35 & 59.94 & 45.55 & 39.42 & 50.02 \\
 & AttnRes & 9.67 & 31.22 & 68.44 & 51.54 & 58.59 & \textbf{32.34} & 87.50 & 23.35 & \textbf{61.41} & 45.81 & 40.95 & 50.11 \\
 & MUDDFormer & 9.38 & 33.38 & 68.61 & 53.91 & 57.70 & \textbf{32.34} & \textbf{89.90} & \textbf{23.96} & 60.58 & 47.05 & 40.53 & 50.80 \\
 & \mthree{} & \textbf{9.31} & \textbf{34.29} & \textbf{68.66} & \textbf{55.25} & 59.93 & 31.57 & 89.60 & 22.27 & 61.25 & \textbf{47.62} & \textbf{41.99} & \textbf{51.24} \\
\cmidrule(l){2-13}
\multirow{3}{*}{1.4B} & Transformer & 8.33 & 38.29 & 71.38 & 58.64 & \textbf{64.65} & 35.49 & 90.40 & 20.74 & 61.04 & 54.56 & 42.97 & 53.82 \\
 & AttnRes & 8.21 & 39.39 & \textbf{71.65} & \textbf{59.04} & 62.71 & \textbf{37.63} & 91.00 & \textbf{24.42} & 62.42 & 54.71 & 42.27 & 54.52 \\
 & \mthree{} & \textbf{8.07} & \textbf{40.97} & 70.84 & 58.88 & 63.76 & 35.92 & \textbf{91.70} & 22.89 & \textbf{64.28} & \textbf{55.92} & \textbf{46.17} & \textbf{55.13} \\
\cmidrule(l){2-13}
\multirow{4}{*}{\begin{tabular}[c]{@{}l@{}}1.8B-\\A0.38B\end{tabular}} & Transformer & 9.73 & 27.38 & 67.63 & 52.17 & \textbf{56.02} & 30.12 & 84.80 & 21.81 & \textbf{61.44} & 44.33 & 35.93 & 48.16 \\
 & AttnRes & 11.12 & 25.07 & 63.82 & 52.96 & 48.91 & 26.54 & 83.80 & 22.89 & 60.37 & 38.74 & 36.84 & 45.99 \\
 & MUDDFormer & 9.97 & 28.90 & 66.59 & 52.49 & 53.91 & 29.78 & 86.40 & \textbf{24.42} & 58.50 & 43.03 & \textbf{39.14} & 48.32 \\
 & \mthree{} & \textbf{9.64} & \textbf{30.33} & \textbf{67.68} & \textbf{53.20} & 54.76 & \textbf{30.29} & \textbf{87.50} & 23.66 & 55.78 & \textbf{44.84} & 37.81 & \textbf{48.58} \\
\cmidrule(l){2-13}
\multirow{2}{*}{\begin{tabular}[c]{@{}l@{}}4.6B-\\A0.89B\end{tabular}} & Transformer & 8.07 & 35.51 & \textbf{72.74} & 56.99 & 63.05 & \textbf{36.26} & \textbf{90.00} & \textbf{20.28} & 55.41 & 54.16 & 42.41 & 52.68 \\
 & \mthree{} & \textbf{8.00} & \textbf{38.00} & 71.06 & \textbf{57.06} & \textbf{64.23} & 36.01 & 89.80 & 19.35 & \textbf{59.85} & \textbf{54.56} & \textbf{43.25} & \textbf{53.32} \\
\bottomrule
\end{tabular}
\caption{Zero-shot downstream accuracy (\%) at 20 training tokens per parameter. PIQA, ARC and HellaSwag use length-normalized accuracy. PPL is the perplexity on the held-out FineWeb-Edu tokens. Bold: best at each size.}
\label{tab:downstream}
\end{table*}

\section{Analysis}
\label{sec:analysis}

\subsection{All three constraints matter}
\label{sec:axes}

Which of the three constraints limit the Transformer?
Table~\ref{tab:axes} ablates them from 100M to 1.4B with the members of the family in \S\ref{sec:granularity}, and with block attention residuals for relaxing only (3).
These lower the loss of the Transformer by 0.047 at 100M but by at most 0.020 from 300M on.
Separating groups as well (multi-group) lowers it by 0.014--0.059, and separating roles instead (multi-role, here MUDDFormer) by 0.028--0.080 up to 700M.
Only \mthree{}, which relaxes all three, goes further, by 0.005--0.013 beyond the better of the two at every size.
At 700M, relaxing (3) alone, (2) and (3), or (1) and (3) recovers 0.005, 0.021 and 0.036 of the 0.044 gain of \mthree{}.
The gains of separating roles and separating groups are not additive, so they partly address the same limitation, but each removes a part that the other leaves in place.
Group-specific mixing helps here even on its own, whereas splitting attention residuals into heads raises their loss \citep{kimi2026attnres}; in \mthree{}, each group has its own signed, unnormalized mixture rather than a share of one softmax over depth.

\begin{table}[t]
\centering
\small
\setlength{\tabcolsep}{3pt}
\begin{tabular}{@{}lccccc@{}}
\toprule
& & & Multi- & Multi- & \\
& Residual & AttnRes$^a$ & group & role$^b$ & \mthree{} \\
(1) Roles  &  &  &  & \checkmark & \checkmark \\
(2) Groups &  &  & \checkmark &  & \checkmark \\
(3) Weights &  & \checkmark & \checkmark & \checkmark & \checkmark \\
\midrule
100M & 2.798 & 2.751 & 2.739 & 2.717 & \textbf{2.713} \\
300M & 2.437 & 2.417 & 2.405 & 2.394 & \textbf{2.383} \\
500M & 2.345 & 2.345 & 2.331 & 2.317 & \textbf{2.304} \\
700M & 2.275 & 2.269 & 2.253 & 2.238 & \textbf{2.231} \\
1.4B & 2.119 & 2.105 & 2.101 & div. & \textbf{2.088} \\
\bottomrule
\end{tabular}
\caption{Ablation of the three constraints: each column relaxes the checked ones. Validation loss at 20 tokens per parameter. $^a$Block attention residuals \citep{kimi2026attnres}, which use a softmax over blocks of layers and, unlike the other columns, are not in the family of \S\ref{sec:granularity}. $^b$MUDDFormer \citep{xiao2025muddformer}, which diverged (div.) at 1.4B.}
\label{tab:axes}
\end{table}

\subsection{How fine should the groups be?}
\label{sec:groups}

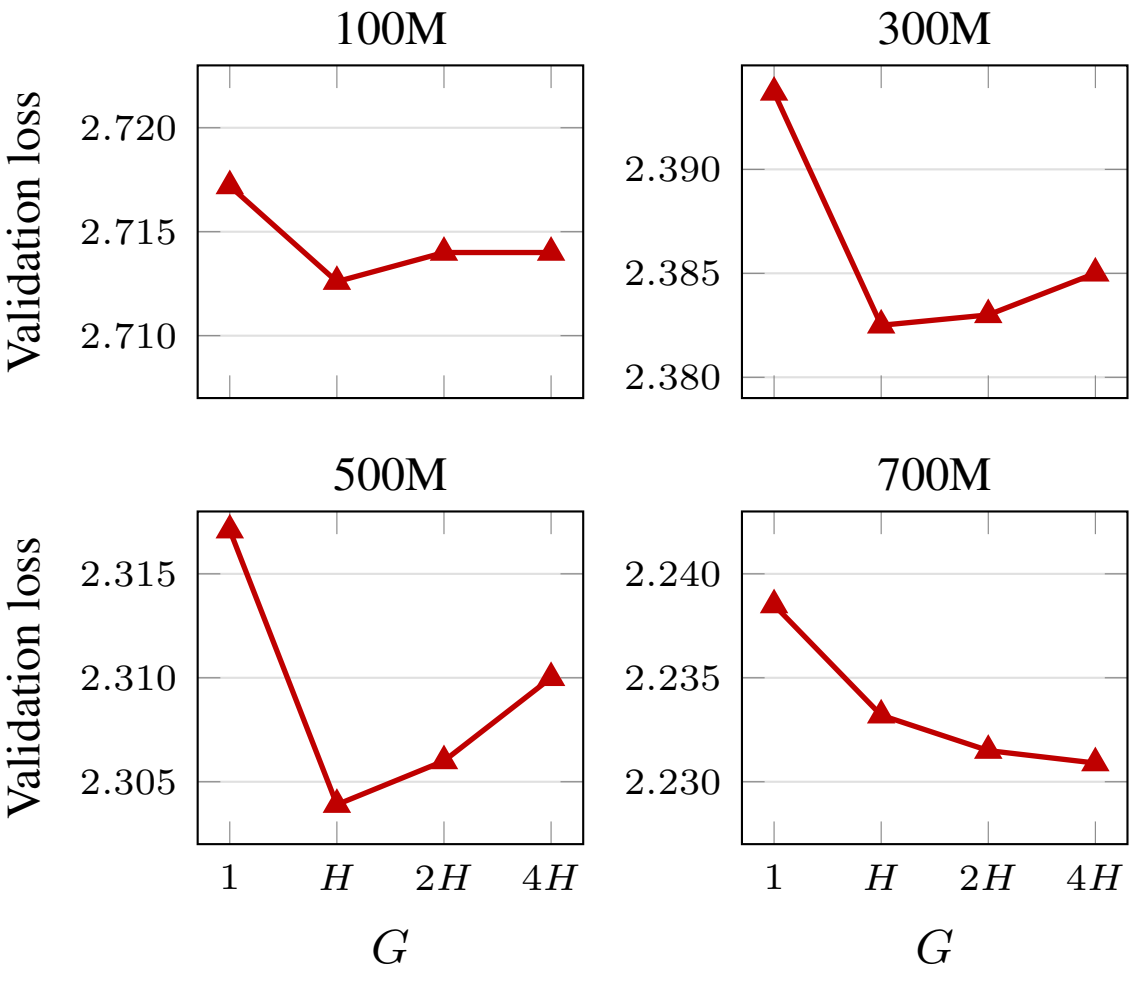
\begin{figure}[t]
\centering
\begin{tikzpicture}
\begin{groupplot}[
  group style={group size=2 by 2, horizontal sep=1.05cm, vertical sep=0.75cm,
    x descriptions at=edge bottom},
  scale only axis, width=2.55cm, height=2.2cm,
  xmin=-0.3, xmax=3.3,
  xtick={0,1,2,3}, xticklabels={$1$,$H$,$2H$,$4H$},
  yticklabel style={/pgf/number format/fixed,/pgf/number format/fixed zerofill,/pgf/number format/precision=3},
  scaled y ticks=false,
  xlabel={$G$},
  label style={font=\small}, tick label style={font=\scriptsize},
  title style={font=\small, yshift=-1.2ex},
  every axis plot/.append style={line width=1pt},
  ymajorgrids, grid style={black!12},
]
\nextgroupplot[title={100M}, ylabel={Validation loss}, ymin=2.707, ymax=2.723, ytick={2.71,2.715,2.72}]
\addplot[cmt, mark=triangle*, mark size=2.1pt] coordinates {(0,2.7172) (1,2.7126) (2,2.714) (3,2.714)};
\nextgroupplot[title={300M}, ymin=2.379, ymax=2.395, ytick={2.38,2.385,2.39}]
\addplot[cmt, mark=triangle*, mark size=2.1pt] coordinates {(0,2.3937) (1,2.3825) (2,2.383) (3,2.385)};
\nextgroupplot[title={500M}, ylabel={Validation loss}, ymin=2.302, ymax=2.318, ytick={2.305,2.31,2.315}]
\addplot[cmt, mark=triangle*, mark size=2.1pt] coordinates {(0,2.3171) (1,2.3039) (2,2.306) (3,2.310)};
\nextgroupplot[title={700M}, ymin=2.227, ymax=2.243, ytick={2.23,2.235,2.24}]
\addplot[cmt, mark=triangle*, mark size=2.1pt] coordinates {(0,2.2385) (1,2.2332) (2,2.2315) (3,2.2309)};
\end{groupplot}
\end{tikzpicture}
\caption{Validation loss with $G$ groups per role, where $H$ is the number of attention heads; $G=1$ is MUDDFormer.}
\label{fig:groups}
\end{figure}

Figure~\ref{fig:groups} varies the number of groups per role, $G\in\{1,H,2H,4H\}$, from 100M to 700M at 20 tokens per parameter; $G=1$ is MUDDFormer.
Groups pay off most at the first split: $G=H$ lowers the loss by 0.005--0.013 at every size, three to nine times what its added parameters would give in a larger Transformer according to the fit (Appendix~\ref{app:scaling}).
Finer groups help only at 700M, by another 0.002, and with $G=H$ there \mformer{} is still 0.005 below MUDDFormer, so the choice of $G$ barely affects our comparisons.

\begin{figure*}[t]
\centering
\includegraphics[width=0.86\textwidth]{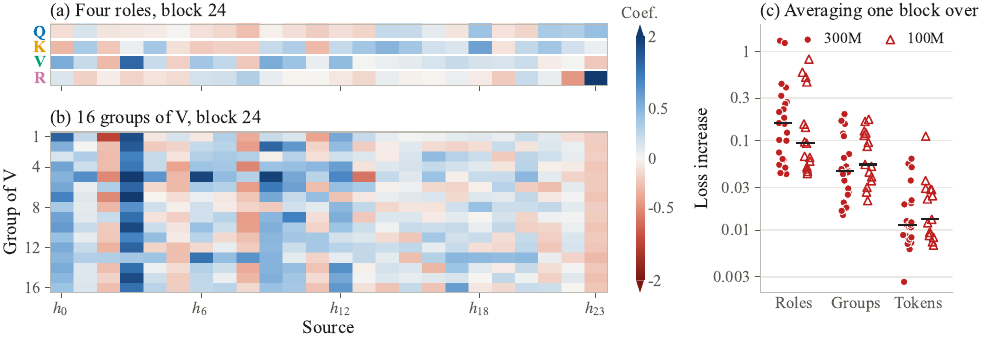}
\caption{Coefficients of separately trained 100M and 300M \mthree{} models ($G=16$), averaged over validation tokens; colors scale as $\mathrm{sign}(a)|a|^{0.65}$. (a), (b): 300M. (c): increase of the validation loss when the coefficients of one block are averaged over roles, groups or tokens; one point per block, bars: medians.}
\label{fig:learned}
\end{figure*}

\subsection{Training stability}
\label{sec:stability}

All models are trained with the same pre-norm recipe and no additional stabilization, and every run from 100M to 700M completed.
At 1.4B, MUDDFormer diverged after 177,788 of 208,021 steps: its loss jumped to 10.37, the loss of a uniform prediction over the vocabulary, its gradient norm fell to zero, and it did not recover.
\mformer{} and attention residuals completed the same run and its decay branches.
MUDDFormer trains its deep variants with PrePostNorm, which normalizes the inputs of the mixing and the mixed outputs \citep{xiao2025muddformer}.
With PrePostNorm, both complete the 1.4B run: MUDDFormer reaches 2.083 and \mformer{} 2.073, so \mformer{} stays ahead by 0.010; PrePostNorm also lowers the loss of \mformer{}, from 2.088.

\subsection{What do roles and groups read?}
\label{sec:learned}

Figure~\ref{fig:learned} shows the coefficients that \mthree{} learns at 300M; Appendix~\ref{app:learned} shows all blocks at both sizes.
The four roles of a block read different layers (a): in the last block, R puts 2.55 on the latest state, Q and K spread over the upper half of the network, and V draws mostly on $h_3$ and the embedding $h_0$ while giving the latest state $-0.15$.
The groups of a role differ as well (b): every group of V reads $h_3$, but with weights from 0.31 to 2.15, and the weights of the groups of K on $h_0$ range from $-0.80$ to $0.52$, so they disagree even in sign.
Reimposing one constraint in a single block, by averaging its coefficients over roles, groups or tokens, raises the loss in every block (c; medians 0.16, 0.046 and 0.011 at 300M).
Averaging over groups costs more than over tokens in all 23 blocks, most of all for V (Appendix~\ref{app:intervene}).

\section{Related Work}
\label{sec:related}

\paragraph{Residual connections.}
Residual connections \citep{he2016deep} and highway networks \citep{srivastava2015training} made deep networks trainable, and in Transformers the residual stream is the channel through which layers communicate \citep{elhage2021mathematical}.
Much work changes how a block writes to this stream, through normalization placement \citep{xiong2020layer,liu2020understanding} or rescaled residual branches \citep{bachlechner2021rezero,wang2022deepnet}.
\mthree{} leaves the write unchanged and changes how each block reads.

\paragraph{Cross-layer connections.}
DenseNet connects every layer to all earlier ones \citep{huang2017densely}, and DLCL \citep{wang2019learning} and DenseFormer \citep{pagliardini2024denseformer} bring learned static weights over earlier layers to Transformers.
Hyper-connections widen the residual stream into several streams with partly token-dependent connections \citep{zhu2025hyperconnections,xie2025mhc}, and frac-connections split the hidden state into fractions \citep{zhu2025frac}; both recombine the current streams rather than read individual earlier layers.
Attention residuals attend over the outputs of earlier layers \citep{kimi2026attnres}, and multi-head and multi-stream variants split this attention by subspace or stream \citep{luo2026mhar,qin2026msar}; all of them form one input per block.
Value residual learning passes the first-layer values to later layers \citep{zhou2025value}.
Closest to our work, DeepCrossAttention \citep{heddes2025deepcrossattention} and MUDDFormer \citep{xiao2025muddformer} form several block inputs from separate mixtures.
DeepCrossAttention combines static channel-wise weights with a token-dependent weight per layer, and MUDDFormer generates token-dependent weights per role; in both, the token-dependent weights are shared by all channels, and \mthree{} makes them specific to the group as well, applying to the depth axis the idea of multi-head attention \citep{vaswani2017attention} and fine-grained experts \citep{dai2024deepseekmoe,krajewski2024scaling}, which split a component into smaller, specialized parts.

\section{Conclusion}
\label{sec:conclusion}

\mthree{} removes the broadcast bottleneck of the residual connection by letting every group of every role read its own token-dependent mixture of earlier layers.
From 100M to 1.4B parameters, \mformer{} matches a Transformer 1.19--1.34$\times$ the size of the one it modifies, improves on MUDDFormer at every size, including 1.4B, where MUDDFormer needs PrePostNorm to converge, and carries over to mixture-of-experts models.
All three constraints matter, and the first split into groups captures most of the gain per parameter.

\section*{Limitations}

Due to computational constraints, our dense models reach 1.4B parameters and our mixture-of-experts models 4.6B; we leave larger models and longer training to future work.
We study decoder-only language models pre-trained on English text, and applying \mthree{} to other architectures, languages and modalities remains to be explored.

\bibliography{references}

\appendix

\section{Derivations}
\label{app:derivations}

This appendix proves the properties of \mthree{} that the main text uses and derives its cost, in the notation of \S\ref{sec:method}.
A \emph{reader} is a pair $i=(r,g)$ of a role $r\in\mathcal{R}$ and a group $g$, so a block has $4G$ readers, and reader $i$ of block~$\ell$ forms $z_{\ell,i}=\sum_{j<\ell}a_{\ell,i,j}\,h^{(g)}_j$.

\subsection{Mixing states and mixing updates}
\label{app:updates}

Let $\Delta_0=h_0$ and $\Delta_k=h_k-h_{k-1}$ for $k\ge1$ be the update of block~$k$, so that $h_j=\sum_{k\le j}\Delta_k$.

\begin{lemma}
\label{lem:updates}
For any $a\in\mathbb{R}^{\ell}$, $\sum_{j<\ell}a_jh_j=\sum_{k<\ell}b_k\Delta_k$ with $b_k=\sum_{j=k}^{\ell-1}a_j$, and $a\mapsto b$ is a bijection of $\mathbb{R}^\ell$.
The residual connection, $a=e_{\ell-1}$, gives $b=\mathbf{1}$, and a convex combination of states gives $1=b_0\ge b_1\ge\dots\ge b_{\ell-1}\ge0$.
\end{lemma}

\begin{proof}
Substitute $h_j=\sum_{k\le j}\Delta_k$ and exchange the sums; the inverse is $a_j=b_j-b_{j+1}$ with $b_\ell:=0$.
For a convex combination, each $b_k$ is a sum of non-negative weights that sum to one.
\end{proof}

The residual connection thus gives the update of every earlier block the weight one, for every token.
A convex mixture, such as a softmax over depth, can only lower the weight of an earlier update, never raise it above one or make it negative, whereas signed, unnormalized coefficients, as in \mthree{}, can give the updates any weights $b\in\mathbb{R}^\ell$.

\subsection{Exact initialization}
\label{app:init}

\begin{proposition}
\label{prop:init}
Let $W_{\ell,2}=0$, let $a^{(0)}_\ell$ be one on $j=\ell-1$ and zero elsewhere for every reader, initialize the output mixer in the same way, and let the scales of $\mathrm{N}_K$ and $\mathrm{N}_V$ equal that of $\mathrm{N}_A$.
Then, for every input and every value of $W_{\ell,1}$, \mformer{} computes the same hidden states and logits as the Transformer with the same remaining parameters.
\end{proposition}

\begin{proof}
Both models share $h_0$.
Suppose they agree on $h_0,\dots,h_{\ell-1}$.
With $W_{\ell,2}=0$, Eq.~\eqref{eq:gen} gives $A_\ell=a^{(0)}_\ell$ whatever $W_{\ell,1}$ is, so $z^{(g)}_{\ell,r}=h^{(g)}_{\ell-1}$ for every reader and $z_{\ell,r}=h_{\ell-1}$.
Since $\mathrm{N}_K$ and $\mathrm{N}_V$ then act as $\mathrm{N}_A$, Eq.~\eqref{eq:m3block} reduces to Eq.~\eqref{eq:block}, and the models agree on $h_\ell$.
By induction they agree on $h_L$, and the output mixer passes $h_L$ to the output layer.
\end{proof}

\subsection{Nesting of the family}
\label{app:nesting}

\begin{proposition}
\label{prop:nest}
Let every member of the family of \S\ref{sec:granularity} use the block of Eq.~\eqref{eq:m3block} and a generator of the form of Eq.~\eqref{eq:gen} whose hidden width is at most $m_\ell$, and let $\mathcal{F}_{\mathrm{res}}$, $\mathcal{F}_{\mathrm{dense}}$, $\mathcal{F}_{\mathrm{role}}$, $\mathcal{F}_{\mathrm{group}}$ and $\mathcal{F}_{\mathrm{M3}}$ be the sets of functions computed by the residual connection, dense, multi-role and multi-group mixing, and \mthree{}, the last two with the same $G$.
Then $\mathcal{F}_{\mathrm{res}}\subseteq\mathcal{F}_{\mathrm{dense}}\subseteq\mathcal{F}_{\mathrm{role}}\subseteq\mathcal{F}_{\mathrm{M3}}$ and $\mathcal{F}_{\mathrm{dense}}\subseteq\mathcal{F}_{\mathrm{group}}\subseteq\mathcal{F}_{\mathrm{M3}}$.
\end{proposition}

\begin{proof}
With $W_{\ell,2}=0$ and $a^{(0)}_\ell$ on the latest state, every member computes the residual connection, as in Proposition~\ref{prop:init}.
A generator of smaller hidden width is a generator of width $m_\ell$ whose extra hidden units have zero output weights, so we can take all hidden widths equal to $m_\ell$.
Given multi-role mixing, whose generator outputs are indexed by $(r,j)$, give \mthree{} the same $W_{\ell,1}$ and set its column $(r,g,j)$ of $W_{\ell,2}$ and entry $(r,g,j)$ of $a^{(0)}_\ell$ to the column and entry $(r,j)$ of multi-role mixing for every $g$; then $a_{\ell,r,g,j}(t)=a_{\ell,r,j}(t)$ for every token, and the two models compute the same function.
Multi-group mixing is embedded in the same way, by repeating its columns and entries $(g,j)$ over the four roles, and dense mixing is embedded in multi-role or multi-group mixing by repeating its columns and entries $j$ over the roles or the groups.
The output mixer is embedded in the same way.
\end{proof}

\mthree{} thus computes every function that a coarser member computes, so its lowest attainable training loss is at most theirs; whether training finds the additional functions is what \S\ref{sec:axes} tests.
With $G=1$, \mthree{} is multi-role mixing, which is our MUDDFormer baseline (\S\ref{sec:setup}).
Attention residuals are not members of the family: their weights are a softmax over the outputs of blocks of layers, computed from all earlier states rather than from $h_{\ell-1}$ alone.

\subsection{Parameters and computation}
\label{app:cost}

Summing the generator and the two extra norms of every block and the output mixer, which has one role and $G$ groups, \mthree{} adds
\begin{equation}
\begin{aligned}
P_{\mathrm{M3}}&=\sum_{\ell=2}^{L}\big[d\,m_\ell+4G\ell\,(m_\ell+1)+2d\big]\\
&\quad+d\,m_{L+1}+G(L+1)(m_{L+1}+1)
\end{aligned}
\label{eq:params}
\end{equation}
parameters, the parameter count of our implementation.
Per token, block~$\ell$ spends $(d+4G\ell)\,m_\ell$ multiply--adds in the generator and $4\ell d$ in Eq.~\eqref{eq:mix}, and the output mixer $(d+G(L+1))\,m_{L+1}+(L+1)d$.
If $m_\ell$ were exactly $4\ell$, the generators would add about $2L^2d+16GL^3/3$ parameters, and relative to the $L(4d^2+3d\,d_{\mathrm{ff}})$ non-embedding parameters of a Transformer with feed-forward width $d_{\mathrm{ff}}$,
\begin{equation}
\frac{P_{\mathrm{M3}}}{N}\approx\frac{2L+16GL^2/(3d)}{4d+3d_{\mathrm{ff}}}=O\Big(\frac{L}{d}+\frac{GL^2}{d^2}\Big),
\label{eq:overhead}
\end{equation}
so the overhead grows with depth relative to width.
In our models, rounding $m_\ell$ up to a multiple of 64 makes $m_\ell$ larger than $4\ell$, so we compute the overhead exactly from Eq.~\eqref{eq:params}; Table~\ref{tab:parambound} lists it for the models we report.

\subsection{Equivalent size and parameter-matched gain}
\label{app:scaling}

We fit $L(N)=E+A\,(N/N_0)^{-\alpha}$ with $N_0=10^8$ to the Transformers trained on $20N$ tokens.
A loss $\ell^\star>E$ is reached by a Transformer of size $N_{\mathrm{eq}}=N_0\big((\ell^\star-E)/A\big)^{-1/\alpha}$, and we report $k=N_{\mathrm{eq}}/N$.
Training on $D=20N$ tokens costs $C\approx6ND=120N^2$ operations, so the Transformer that matches the loss uses $k^2$ times the compute of a Transformer of size $N$.

To compare the gain of \mthree{} with what its extra parameters would give if they were spent on a larger Transformer, we evaluate the fit at $(1+\delta)N$, where $\delta=P_{\mathrm{M3}}/N$:
\begin{equation}
\begin{aligned}
&L(N)-L\big((1+\delta)N\big)\\
&\qquad=A\,(N/N_0)^{-\alpha}\big[1-(1+\delta)^{-\alpha}\big].
\end{aligned}
\label{eq:parambound}
\end{equation}
This is an estimate under the fitted Transformer curve, not a bound on what parameters in another form could contribute.
Table~\ref{tab:parambound} compares it with the observed gain.

\begin{table}[h]
\centering
\small
\setlength{\tabcolsep}{4pt}
\begin{tabular}{lcccccc}
\toprule
Size & $G$ & Params & MACs & Fit & Observed \\
\midrule
100M & $H=10$ & 1.44\% & 1.85\% & 0.0043 & 0.0850 \\
300M & $H=16$ & 1.40\% & 1.78\% & 0.0033 & 0.0542 \\
500M & $H=18$ & 1.40\% & 1.80\% & 0.0031 & 0.0412 \\
700M & $4H=80$ & 3.86\% & 4.25\% & 0.0081 & 0.0438 \\
1.4B & $H=26$ & 1.52\% & 1.91\% & 0.0029 & 0.0308 \\
\bottomrule
\end{tabular}
\caption{Overhead of \mthree{} at the reported $G$, relative to the non-embedding parameters and per-token weight multiply--adds (MACs) of the Transformer, and loss reduction at $20N$: from the extra parameters by Eq.~\eqref{eq:parambound} (Fit) and observed over the Transformer.}
\label{tab:parambound}
\end{table}

\section{Training Details}
\label{app:training}

\begin{table}[h]
\centering
\small
\begin{tabular}{@{}l>{\raggedright\arraybackslash}p{4.3cm}@{}}
\toprule
Data & FineWeb-Edu (sample-350BT), fixed subset \\
Tokenizer & TinyLlama SentencePiece (32,000) \\
Sequence length & 2,048 \\
Global batch & 64 sequences (131,072 tokens) \\
Optimizer & AdamW, $\beta_1=0.9$, $\beta_2=0.95$, $\epsilon=10^{-8}$, weight decay 0.1 \\
Gradient clipping & global norm 1.0 \\
Schedule & WSD; linear warmup over 1\% of the $20N$ budget; $1-\sqrt{\cdot}$ decay to 10\% of the peak over the last 20\% of each budget \\
Peak learning rate & $(7.6, 4.55, 3.2, 2.6, 1.9)\times10^{-4}$ for 100M, 300M, 500M, 700M, 1.4B \\
Budgets & $5N$, $10N$, $20N$ tokens; $N$ is the non-embedding parameter count of the Transformer \\
Validation & 99,876,864 tokens \\
Precision & BF16 computation; FP32 master parameters and gradient reduction \\
Hardware & NVIDIA H200 \\
\bottomrule
\end{tabular}
\caption{Training hyperparameters of the dense models, shared by all methods.}
\label{tab:hparams}
\end{table}

Table~\ref{tab:hparams} lists the training hyperparameters of the dense models.
The dense Transformer is a pre-norm decoder with RMSNorm ($\epsilon=10^{-5}$), SwiGLU feed-forward layers, rotary position embeddings with base 10,000, and tied input and output embeddings.
RMSNorm scales start at one.
The query, key and value projections and the gate projection of the feed-forward layer are drawn from a truncated normal distribution with standard deviation 0.02, and the attention output projection and the up and down projections of the feed-forward layer from one with standard deviation $0.02/\sqrt{2\ell}$ in block $\ell$, counted from one.
The tied embedding is drawn from a normal distribution with standard deviation $d^{-1/2}$, truncated at $\pm3d^{-1/2}$.
Each backbone parameter is seeded by its name, so all methods start from the same backbone; \S\ref{sec:weights} gives the initialization of the \mthree{} mixers.
The 1.4B main run has 208,021 steps and the 700M main run 98,013 steps.
Table~\ref{tab:moearch} gives the mixture-of-experts architectures of \S\ref{sec:moe}; parameter counts are those of the Transformer baselines, and MUDDFormer and \mthree{} add their mixers.

\begin{table}[h]
\centering
\small
\setlength{\tabcolsep}{4pt}
\begin{tabular}{lcc}
\toprule
& Small & Large \\
\midrule
Blocks $L$ & 24 & 28 \\
Width $d$ & 1024 & 1536 \\
Attention heads (MHA) & 16 & 24 \\
Head dimension & 64 & 64 \\
Experts per layer & 64 & 64 \\
Experts per token & 8 & 8 \\
Expert width & 352 & 512 \\
Expert activation & SwiGLU & SwiGLU \\
Shared experts & none & none \\
Vocabulary & 32,000 & 32,000 \\
Embedding and output & untied & untied \\
Sequence length & 2,048 & 2,048 \\
Total / active params & 1.83B / 0.38B & 4.59B / 0.89B \\
\mthree{} groups $G$ & 64 & 96 \\
\bottomrule
\end{tabular}
\caption{Mixture-of-experts architectures. Every feed-forward layer is a mixture-of-experts layer, and experts are not shared across layers.}
\label{tab:moearch}
\end{table}

\section{Full Results}
\label{app:full}

\begin{table*}[t]
\centering
\scriptsize
\setlength{\tabcolsep}{4.2pt}
\begin{tabular}{@{}llccccccccccc@{}}
\toprule
Size & Model & LAMBADA & PIQA & WinoGrande & ARC-E & ARC-C & SciQ & LogiQA & BoolQ & HellaSwag & RACE-M & Avg \\
\midrule
\multirow{4}{*}{700M} & Transformer & 29.90 & 68.72 & 55.33 & 65.70 & 32.85 & 90.20 & 22.43 & \textbf{62.26} & 45.71 & 41.50 & 51.46 \\
 & AttnRes & 30.04 & 67.41 & 52.80 & 66.37 & 34.04 & 90.60 & 20.58 & 56.73 & 45.37 & 41.57 & 50.55 \\
 & MUDDFormer & 30.56 & 68.66 & 54.06 & 68.48 & 34.47 & 90.40 & \textbf{25.04} & 58.65 & 47.06 & 42.20 & 51.96 \\
 & \mthree{} & \textbf{33.71} & \textbf{69.04} & \textbf{55.96} & \textbf{68.69} & \textbf{34.90} & \textbf{91.80} & 23.35 & 53.15 & \textbf{47.84} & \textbf{43.38} & \textbf{52.18} \\
\cmidrule(l){2-12}
\multirow{3}{*}{1.4B} & Transformer & 36.25 & \textbf{72.25} & 59.35 & 72.39 & 39.16 & 93.90 & 22.43 & 62.78 & 54.67 & 45.26 & 55.85 \\
 & AttnRes & 40.40 & 71.55 & 58.64 & 72.35 & \textbf{40.36} & 93.40 & \textbf{23.50} & \textbf{62.97} & 54.72 & 44.85 & 56.27 \\
 & \mthree{} & \textbf{40.42} & 71.65 & \textbf{61.17} & \textbf{72.90} & 39.51 & \textbf{94.60} & 21.81 & 59.69 & \textbf{56.28} & \textbf{47.14} & \textbf{56.52} \\
\cmidrule(l){2-12}
\multirow{4}{*}{\begin{tabular}[c]{@{}l@{}}1.8B-\\A0.38B\end{tabular}} & Transformer & 26.14 & \textbf{69.04} & 51.85 & \textbf{65.78} & \textbf{33.28} & 88.60 & 25.81 & 56.82 & 44.19 & 36.00 & 49.75 \\
 & AttnRes & 24.65 & 64.64 & 53.59 & 56.40 & 28.24 & 85.80 & 22.58 & 43.46 & 38.07 & 36.00 & 45.34 \\
 & MUDDFormer & 24.53 & 67.90 & \textbf{54.30} & 63.05 & 32.76 & \textbf{89.20} & 25.81 & \textbf{59.36} & 42.87 & \textbf{39.35} & \textbf{49.91} \\
 & \mthree{} & \textbf{26.57} & 67.57 & 51.38 & 64.65 & 32.34 & 89.10 & \textbf{26.73} & 53.06 & \textbf{44.84} & 38.65 & 49.49 \\
\cmidrule(l){2-12}
\multirow{2}{*}{\begin{tabular}[c]{@{}l@{}}4.6B-\\A0.89B\end{tabular}} & Transformer & \textbf{36.99} & \textbf{73.01} & \textbf{56.99} & \textbf{71.97} & \textbf{40.27} & 92.30 & 21.51 & 60.24 & 54.24 & 43.31 & 55.08 \\
 & \mthree{} & 36.02 & 71.60 & 56.67 & \textbf{71.97} & 38.74 & \textbf{93.30} & \textbf{23.96} & \textbf{61.19} & \textbf{55.14} & \textbf{46.03} & \textbf{55.46} \\
\bottomrule
\end{tabular}
\caption{Five-shot downstream accuracy (\%) at 20 training tokens per parameter, with the metrics of Table~\ref{tab:downstream}. Bold: best at each size.}
\label{tab:downstream5}
\end{table*}

Table~\ref{tab:full} lists all runs of the main setup.
Table~\ref{tab:full700} adds all 700M runs.
Table~\ref{tab:downstream5} gives five-shot downstream accuracy; the five examples are fixed across models, RACE-M is scored per question, and prompts are not truncated.
We compare methods at $20N$ in the main text; with $5N$ tokens, MUDDFormer is slightly ahead of \mthree{} at 100M (by 0.003) and of \mthree{} with $G=H$ at 700M (by 0.012).

\begin{table}[h]
\centering
\small
\setlength{\tabcolsep}{4pt}
\resizebox{\columnwidth}{!}{%
\begin{tabular}{llrcccc}
\toprule
Size & $D$ & Tokens & Transf. & AttnRes & MUDDFormer & \mthree{} \\
\midrule
100M & $5N$  & 0.41B & 3.0899 & 3.0447 & 2.9780 & 2.9811 \\
     & $10N$ & 0.81B & 2.9209 & 2.8743 & 2.8289 & 2.8273 \\
     & $20N$ & 1.63B & 2.7976 & 2.7507 & 2.7172 & 2.7126 \\
\midrule
300M & $5N$  & 1.54B & 2.6395 & 2.6183 & 2.5836 & 2.5724 \\
     & $10N$ & 3.08B & 2.5243 & 2.5043 & 2.4774 & 2.4669 \\
     & $20N$ & 6.17B & 2.4367 & 2.4168 & 2.3937 & 2.3825 \\
\midrule
500M & $5N$  & 2.29B & 2.5368 & 2.5292 & 2.5043 & 2.4878 \\
     & $10N$ & 4.58B & 2.4278 & 2.4257 & 2.4002 & 2.3853 \\
     & $20N$ & 9.17B & 2.3452 & 2.3447 & 2.3171 & 2.3039 \\
\midrule
700M & $5N$  & 3.21B  & 2.4566 & 2.4455 & 2.4139 & 2.4055 \\
     & $10N$ & 6.42B  & 2.3534 & 2.3469 & 2.3159 & 2.3087 \\
     & $20N$ & 12.85B & 2.2747 & 2.2692 & 2.2385 & 2.2309 \\
\midrule
1.4B & $5N$  & 6.82B  & 2.2763 & 2.2594 & --       & 2.2428 \\
     & $10N$ & 13.63B & 2.1870 & 2.1732 & --       & 2.1556 \\
     & $20N$ & 27.27B & 2.1193 & 2.1053 & diverged & 2.0884 \\
\bottomrule
\end{tabular}}
\caption{Validation loss of all runs in the main setup. \mthree{} uses the best $G$ at each size (Table~\ref{tab:parambound}). Attention residuals use blocks of four layers. Multi-group mixing (Table~\ref{tab:axes}) is trained at $20N$ only, except at 700M (Table~\ref{tab:full700}).}
\label{tab:full}
\end{table}

\begin{table}[h]
\centering
\small
\setlength{\tabcolsep}{5pt}
\begin{tabular}{lccc}
\toprule
700M & $5N$ & $10N$ & $20N$ \\
\midrule
Transformer & 2.4566 & 2.3534 & 2.2747 \\
Attention residuals & 2.4455 & 2.3469 & 2.2692 \\
Multi-group, $G=4H$ & 2.4349 & 2.3316 & 2.2533 \\
MUDDFormer & 2.4139 & 2.3159 & 2.2385 \\
\midrule
\mthree{}, $G=H$ & 2.4259 & 2.3108 & 2.2332 \\
\mthree{}, $G=2H$ & \textbf{2.4032} & 2.3099 & 2.2315 \\
\mthree{}, $G=4H$ & 2.4055 & \textbf{2.3087} & \textbf{2.2309} \\
\bottomrule
\end{tabular}
\caption{All 700M runs of the main setup ($H=20$).}
\label{tab:full700}
\end{table}

\section{Pseudocode}
\label{app:code}

Figure~\ref{fig:code} gives a PyTorch-style reference implementation of one \mformer{} block; our fused kernel computes the same mixture in a single pass over the history.

\begin{figure}[!h]
\centering
\begin{minipage}{\columnwidth}
\footnotesize
\begin{verbatim}
# H: [h_0, ..., h_{l-1}], each of shape [T, d]
# returns h_l
def m3_block(H, blk, G):
    l, (T, d) = len(H), H[-1].shape
    # token-conditioned coefficients a_{l,r,g,j}
    x = rms_norm(H[-1])          # no learned scale
    A = blk.W2(gelu(blk.W1(x))) + blk.a0
    A = A.view(T, 4, G, l)       # roles Q, K, V, R
    # split every state into G groups
    S = torch.stack(H, -1).view(T, G, d // G, l)
    Z = torch.einsum('trgj,tgcj->trgc', A, S)
    zq, zk, zv, zr = Z.reshape(T, 4, d).unbind(1)
    u = zr + blk.attn(blk.na(zq), blk.nk(zk),
                      blk.nv(zv))
    return u + blk.ffn(blk.nf(u))

# init: W2 = 0; a0 = 1 on j = l-1, 0 elsewhere
# -> every role reads h_{l-1} (standard block)
\end{verbatim}
\end{minipage}
\caption{PyTorch-style pseudocode of an \mformer{} block. \texttt{blk.attn} includes the query, key, value and output projections.}
\label{fig:code}
\end{figure}

\section{Learned Coefficients}
\label{app:learned}

\begin{figure*}[h]
\centering
\includegraphics[width=\textwidth]{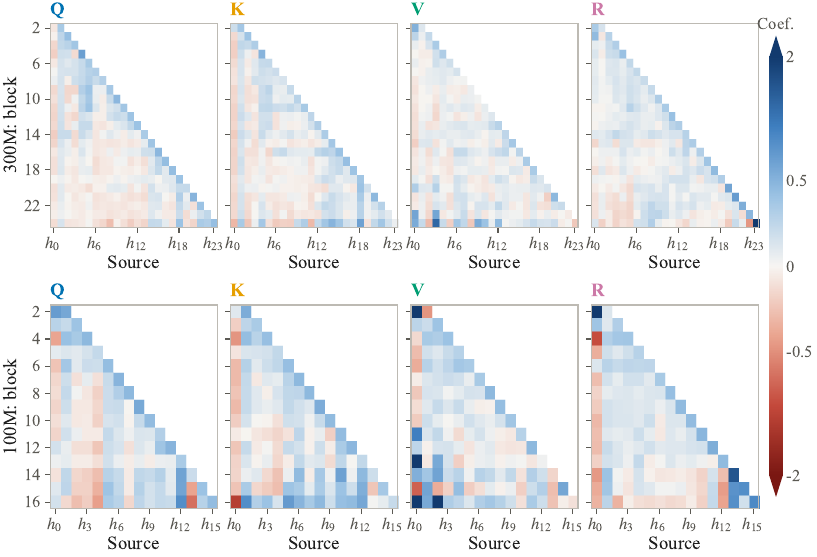}
\caption{Coefficients of every role on every earlier state in every block, averaged over groups and validation tokens, with the colors of Figure~\ref{fig:learned}. Top: 300M; bottom: 100M.}
\label{fig:learnedroles}
\end{figure*}

\begin{figure*}[h]
\centering
\includegraphics[width=\textwidth]{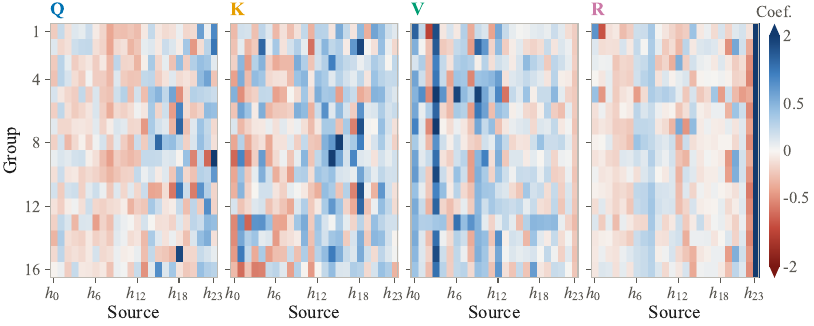}
\caption{Coefficients of the 16 groups of every role in the last block of the 300M model, averaged over validation tokens, with the colors of Figure~\ref{fig:learned}.}
\label{fig:learnedgroups}
\end{figure*}

\begin{figure*}[h]
\centering
\includegraphics[width=\textwidth]{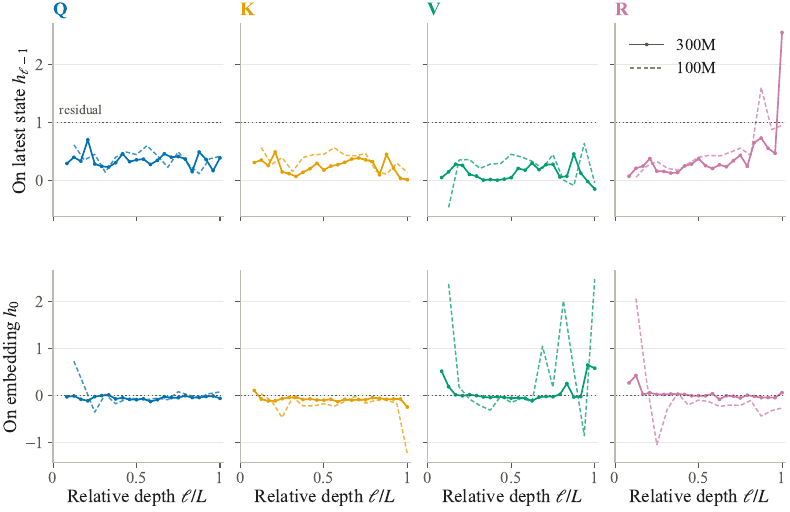}
\caption{Weight of every role on the latest state (a) and on the embedding (b), averaged over groups and validation tokens, against relative depth.}
\label{fig:learnedtrends}
\end{figure*}

Figures~\ref{fig:learnedroles}--\ref{fig:learnedtrends} extend Figure~\ref{fig:learned} with the same models ($G=16$).
Figure~\ref{fig:learnedroles} shows the group-averaged coefficients of every role in every block, Figure~\ref{fig:learnedgroups} all 16 groups of every role in the last block at 300M, and Figure~\ref{fig:learnedtrends} the weights on the latest state and on the embedding through depth.
Through the middle of the network, every role gives the latest state at most 0.73, far below the 1 of the residual connection; only R puts more than 1 on it, and only in the last blocks; V reads the embedding in the first blocks and again in the last third of the network, and K subtracts it in the last block.
In the last block, the groups of R agree on the latest state, whereas the groups of Q and K read different layers.

\section{Interventions on the Coefficients}
\label{app:intervene}

\begin{figure*}[h]
\centering
\includegraphics[width=\textwidth]{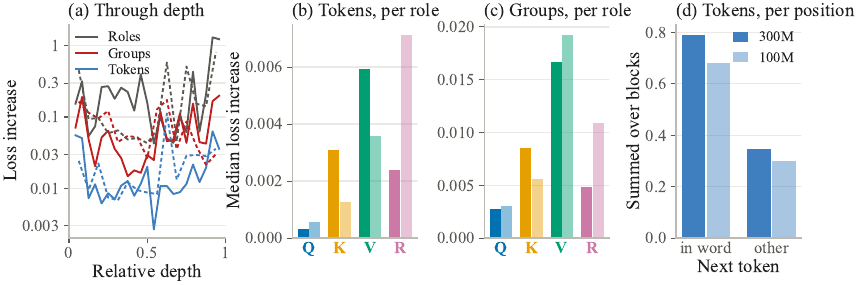}
\caption{Interventions of Figure~\ref{fig:learned}c. (a) Loss increase against relative depth; solid: 300M, dashed: 100M. (b), (c) Median over blocks of the loss increase when only one role is averaged over tokens (b) or over its groups (c). (d) Loss increase from averaging over tokens, summed over blocks, at positions whose next token continues the current word (in word) and at other positions. Dark: 300M; light: 100M.}
\label{fig:intervene}
\end{figure*}

Figure~\ref{fig:intervene} details the interventions of Figure~\ref{fig:learned}c.
We run the models of Figure~\ref{fig:learned} on 128 validation sequences of 2048 tokens and replace the coefficients of one block by an average: over its four roles, over the groups of each role, or over tokens, using the token means on 64 other validation sequences.
Every sequence sees every intervention, and each intervention of Figure~\ref{fig:learned}c raises the loss by at least 5.9 standard errors over the sequences.
Averaging over the four roles, which mixes the residual input with the normalized query, key and value inputs, costs most in the last blocks (a).
Averaged one role at a time, the groups matter most for V at both sizes, whose groups read $h_3$ with weights from 0.31 to 2.15 in the last block (Figure~\ref{fig:learned}b), and the token matters least for Q (b, c).
The dependence on the token is concentrated: 10\% of the tokens carry 92\% of the loss increase summed over blocks at 300M and 87\% at 100M, and positions whose next token continues the current word lose 2.3 times as much as other positions at both sizes (d), mostly through R.
Applied to all blocks at once, averaging over tokens, groups and roles raises the loss by 1.2, 6.2 and 8.4 at 300M.

\end{document}